\documentclass[journal,twoside,web]{ieeecolor}
\usepackage{jsen}
\usepackage{cite}
\usepackage{amsmath,amssymb,amsfonts}
\usepackage{algorithmic}
\usepackage{graphicx}
\usepackage{textcomp}
\usepackage{wrapfig}

\usepackage{pifont}
\usepackage{caption}
\usepackage{hyperref}
\hypersetup{hidelinks}

\def\BibTeX{{\rm B\kern-.05em{\sc i\kern-.025em b}\kern-.08em
    T\kern-.1667em\lower.7ex\hbox{E}\kern-.125emX}}
\definecolor{abstractbg}{rgb}{0.89804,0.94510,0.83137}
\begin{document}
\title{Anomaly Detection for Industrial Applications, Its Challenges, Solutions, and Future Directions: A Review}
\author{Abdelrahman Alzarooni, Ehtesham Iqbal, Samee Ullah Khan, \IEEEmembership{Member, IEEE}, Sajid Javed, Brain Moyo, Yusra Abdulrahman, \IEEEmembership{Member, IEEE}
\thanks{This work is funded by Sanad Aerotech and supported by the Advanced Research and Innovation Center (ARIC), which is jointly funded by Aerospace Holding Company LLC, a wholly-owned subsidiary of Mubadala Investment Company PJSC and Khalifa University for Science and Technology. The corresponding author is Dr. Yusra Abdulrahman (e-mail: \href{mailto:yusra.abdulrahman@ku.ac.ae}{yusra.abdulrahman@ku.ac.ae}).}
\thanks{Abdelrahman Alzarooni and Yusra Abdulrahman are with the Department of Aerospace Engineering and also with the Advanced Research and Innovation Center, Khalifa University of Science and Technology, Abu Dhabi, UAE, (e-mail: \href{mailto:100064457@ku.ac.ae}{100064457@ku.ac.ae}).}
\thanks{Ehtesham Iqbal and Samee Ullah are with the Advanced Research and Innovation Center, Khalifa University of Science and Technology, Abu Dhabi, UAE (e-mail: \href{mailto:ehtesham.iqbal@ku.ac.ae}{ehtesham.iqbal@ku.ac.ae}); (e-mail: \href{mailto:samee.khan@ku.ac.ae}{samee.khan@ku.ac.ae}).}
\thanks{Sajid Javed is with the Department of Computer Science, Khalifa University of Science and Technology, Abu Dhabi, UAE (e-mail: \href{mailto:sajid.javed@ku.ac.ae}{sajid.javed@ku.ac.ae}).}
\thanks{Brain Moyo is with the Research \& Development Program, Sanad Aerotech, Abu Dhabi, UAE (e-mail: \href{mailto:bmoyo@sanad.ae}{bmoyo@sanad.ae}).}}
\IEEEtitleabstractindextext{%
\fcolorbox{abstractbg}{abstractbg}{%
\begin{minipage}{\textwidth}%
\begin{wrapfigure}[22]{r}{3in}%
\includegraphics[width=2.9in]{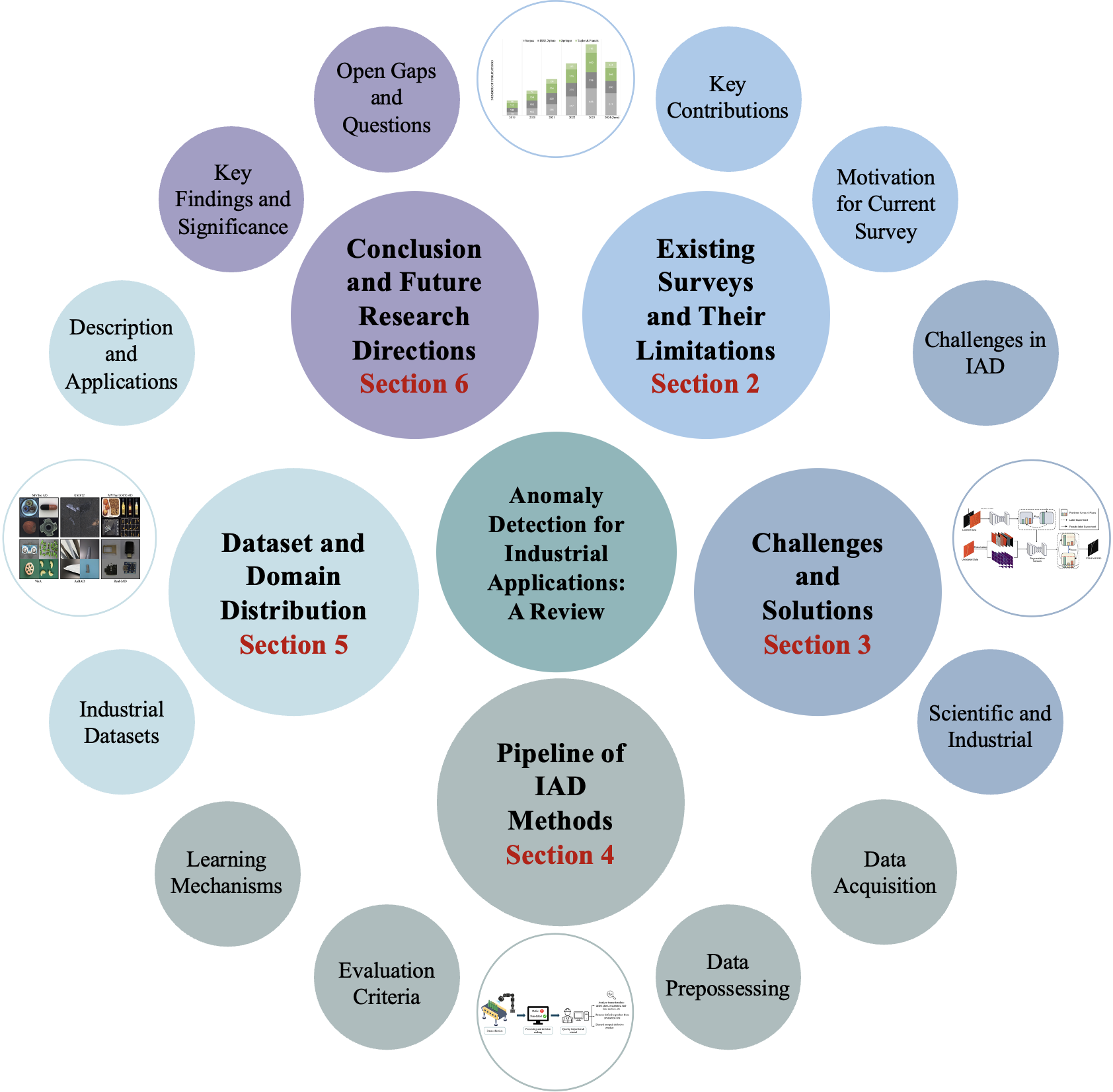}%
\end{wrapfigure}%
\begin{abstract}
Anomaly detection from images captured using camera sensors is one of the mainstream applications at the industrial level. Particularly, it maintains the quality and optimizes the efficiency in production processes across diverse industrial tasks, including advanced manufacturing and aerospace engineering. Traditional anomaly detection workflow is based on a manual inspection by human operators, which is a tedious task. Advances in intelligent automated inspection systems have revolutionized the Industrial Anomaly Detection (IAD) process. Recent vision-based approaches can automatically extract, process, and interpret features using computer vision and align with the goals of automation in industrial operations. 
In light of the shift in inspection methodologies, this survey reviews studies published since 2019, with a specific focus on vision-based anomaly detection. The components of an IAD pipeline that are overlooked in existing surveys are presented, including areas related to data acquisition, preprocessing, learning mechanisms, and evaluation. In addition to the collected publications, several scientific and industry-related challenges and their perspective solutions are highlighted. Popular and relevant industrial datasets are also summarized, providing further insight into inspection applications. Finally, future directions of vision-based IAD are discussed, offering researchers insight into the state-of-the-art of industrial inspection.
\end{abstract}

\end{minipage}}}

\maketitle

\section{Introduction}
\begin{figure*}[h]
\centerline{\includegraphics[width=\textwidth]{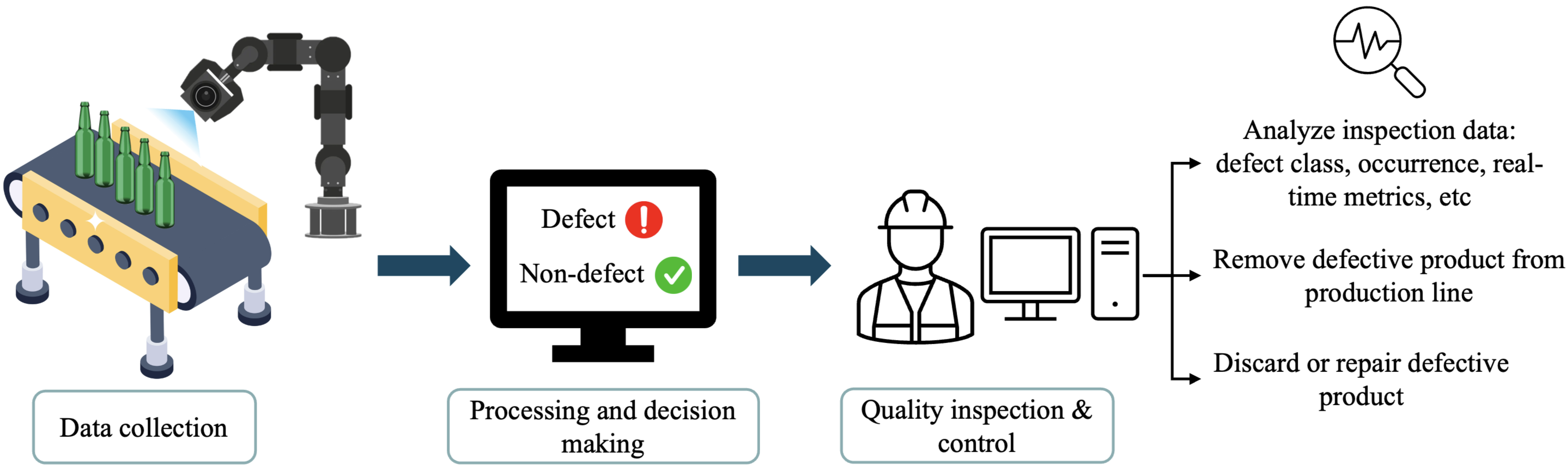}}
\caption{Automated inspection paradigm in a general manufacturing setting. Showcases the data collection, decision making, and quality control process.\label{fig1}}
\end{figure*}
\IEEEPARstart{T}{he} capability in being able to detect anomalies in manufacturing and industrial settings is of high importance, as it ensures that the production process remains controlled and operates as intended \cite{articlerev}. Inspection is especially important during the production of goods, maintaining standards and ensuring consistency throughout the process. For instance, defects present on the surface of a product not only affect its physical appearance but can also degrade its overall performance \cite{articlezen}. Lacking the proper means of quality control in industrial manufacturing results in defects, ranging from minor imperfections to significant structural problems, to be overlooked and subsequently cause product liability and possible injuries. These defects are not localized to a single industry and can be found in additive-manufactured components, Printed Circuit Boards (PCB), and pharmaceutical tablets, among others. Traditionally, defects are identified by manual inspection, which is based on human judgment. Such methods require human inspectors to have significant experience or undergo specialized training to detect defects accurately \cite{9900177}. However, such types of inspection are a highly tedious and time-consuming task, performed by operators who are inconsistent, prone to error, unreliable, and have different personal judgments based on the individual’s risk tolerance. Additionally, the advancement towards the next industrial revolution has given more reasons to explore alternative solutions. The Industry 4.0 revolution, for instance, aligns with improving inspection capabilities Other similar strategies have been proposed, such as “Made in China 2025” \cite{LI201866}, “Society 5.0” in Japan \cite{salgues2018society}, and the “Industrial Internet Consortium” in the United States \cite{QIN20201}. The main aim of these approaches is to transform the manufacturing industry by implementing technologically advanced applications to improve performance, quality, and efficiency. This revolutionary development takes shape by incorporating artificial intelligence, advanced robotics, and the Internet of Things (IoT) \cite{8843824} into the manufacturing domain. Consequently, automated defect detection systems have been developed to mimic the manual inspection process, addressing these limitations and improving efficiencies. Fig. \ref{fig1} illustrates an application of automated inspection in a manufacturing setting. The general process includes imaging the manufactured goods, which can travel on a conveyor belt. The imaging medium can vary depending on the specific application. Images are then sent to a decision station, where defects are identified. Finally, the output data are integrated into the quality control center to gain insight into defects and make informed decisions about the manufacturing process.

Efficient identification and classification of industrial product defects is a key issue that is continuously being examined in the literature. Research into higher detection automation is important due to the complex and varied factors that influence defect formation. Some of these challenges are found in real-world industrial settings. For instance, the surfaces of manufactured products are often exposed to noise introduced during the manufacturing process. Alterations to a surface's appearance make it difficult to distinguish the noise from real defects. In addition, defects can drastically vary in size, placement, texture, and color which further adds complexity to the detection process. Moreover, since some objects of interest in an image are usually small, less than satisfactory texture information can be extracted \cite{ZHENG2024108170}. Nevertheless, there is significant academic interest in monitoring and control systems that aim to further inspection technologies. Progress towards these areas is fueled by the potential of lower production costs, increased yields, and the implementation of smart manufacturing \cite{doi:10.1080}. The incorporation of such advanced control systems is a key enabling technology for the fourth industrial revolution, especially in the quality control of manufactured products \cite{7064655}. Consequently, learning-based algorithms have become an important consideration in achieving an automated inspection system.

\subsection{Learning Algorithms}
Vision-based defect detection has become one of the key technologies used in intelligent manufacturing \cite{articlezen}. These techniques include Machine Learning (ML) and Deep Learning (DL) algorithms, which analyze data and learn to make accurate predictions for future tasks \cite{8948233}. ML can be defined as the set of methods that automatically detect patterns in data and use them for decision-making under uncertainty \cite{BERTOLINI2021114820}. Despite this, DL, a specialized subset of ML, incorporates artificial neural networks that can learn patterns and extract higher-level features from data. In recent years, with the advancement of AI technology, ML and DL have been widely adopted for detecting defects in industrial domains \cite{10113226}. However, another factor which influneces these algorithms is the level of supervision applied during training. Different levels of supervision, including Supervised Learning (SL), Unsupervised Learning (USL), and Semi-Supervised Learning (SSL), each shape how models learn from data and address challenges.

\subsubsection{Supervised Learning}
\begin{figure*}[t]
\centerline{\includegraphics[width=\linewidth]{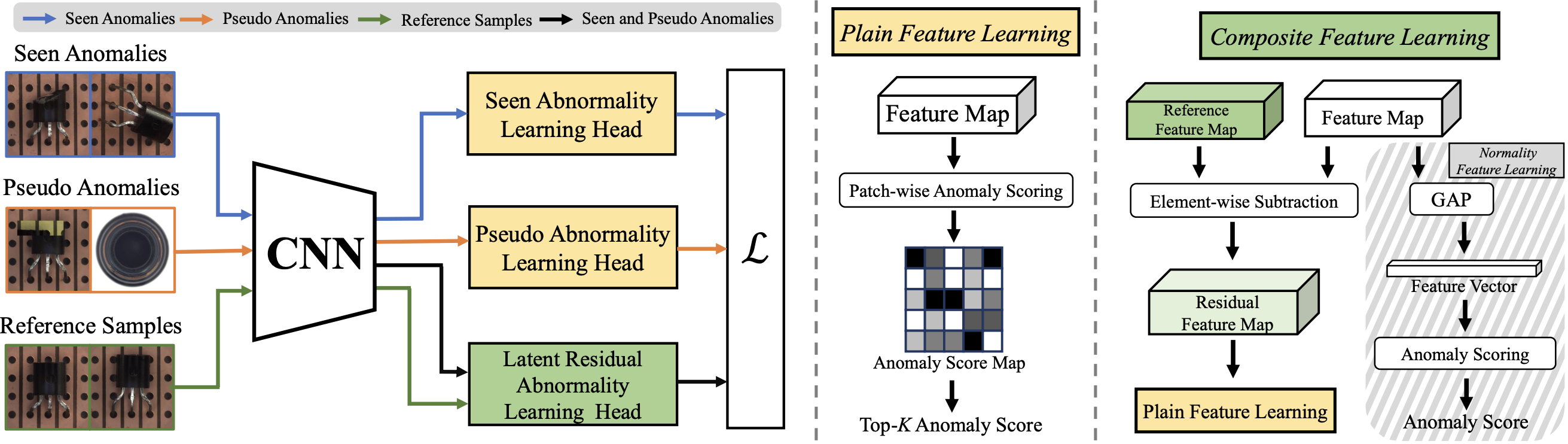}}
\caption{An overview of a supervised anomaly detection framework that learns three disentangled types of abnormalities \cite{ding2022catching}.
Image samples are first fed into a neural network, which extracts feature maps representing key characteristics of the input. Features are then passed to a classifier that assigns labels (e.g., normal or defective) or identifies specific abnormalities based on the feature patterns.\label{fig2}}

\end{figure*}

SL classifies inputs as defective or non-defective by using labeled training data. Such a method can achieve high detection rates due to the availability of labeled data. However, it is not an absolute solution as it faces challenges with data imbalance. An example of such a supervised ML strategy is found in the inspection of additively manufactured components using powder bed fusion \cite{GOBERT2018517}. The authors of this paper trained a Support Vector Machine (SVM) classifier to detect discontinuities using labeled ground truth data. Images were extracted from post-build CT scans and were manually labeled by a skilled operator. Detection accuracies greater than 80\% were achieved following cross-validation experiments. Another SL technique is illustrated in Fig. \ref{fig2}. The authors of this paper proposed a model to tackle open-set supervised anomaly detection \cite{ding2022catching}. The goal was to detect both seen and unseen anomalies by learning disentangled portrayals of irregularities. The model was tested on five industrial defect detection datasets and outperformed other competitive models by as much as 5\% AUC.

\subsubsection{Unsupervised Learning}
\begin{figure}[b!]
\centerline{\includegraphics[width=0.5\textwidth]{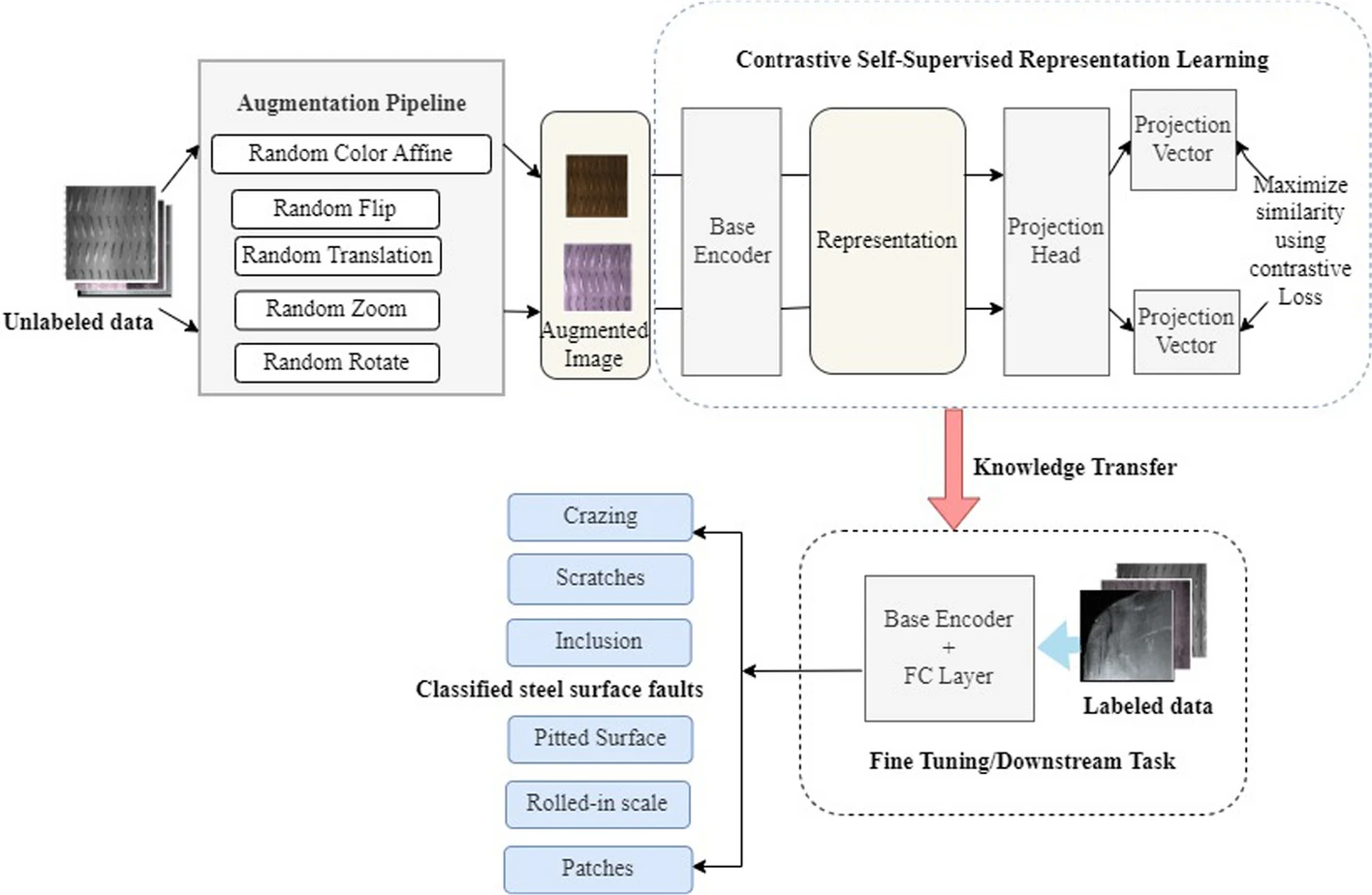}}
\caption{Diagram of a self-supervised model's flow, from image input to the classification of steel surface defects \cite{zabin2023contrastive}.\label{fig3}}
\end{figure}

If labeled data are unavailable, USL can be employed. This method operates by incorporating a dataset that contains solely unlabeled samples, with only a few defective ones. A One-Class Support Vector Machine (OCSVM) was proposed for defect detection in electron microscopy images of crystal structures \cite{artun}. Based on USL, the model achieved a balanced accuracy of 98\%. It demonstrated applicability to various defects while overcoming the need for manual annotations of complex crystals. Self-supervised learning is a subset of USL that also utilizes unlabeled data. The latter is used for anomaly detection, which does not require a loss function, whereas the former is employed in regression and classification tasks similar to those in supervised learning. An example of a self-supervised method is proposed by \cite{zabin2023contrastive}. The diagram in Fig. \ref{fig3} outlines the workflow of this model. The framework was applied to the inspection of metal surfaces, demonstrating a classification accuracy of 97.78\%.

\subsubsection{Semi-supervised Learning}
\begin{figure}[t!]
\centerline{\includegraphics[width=3.5in]{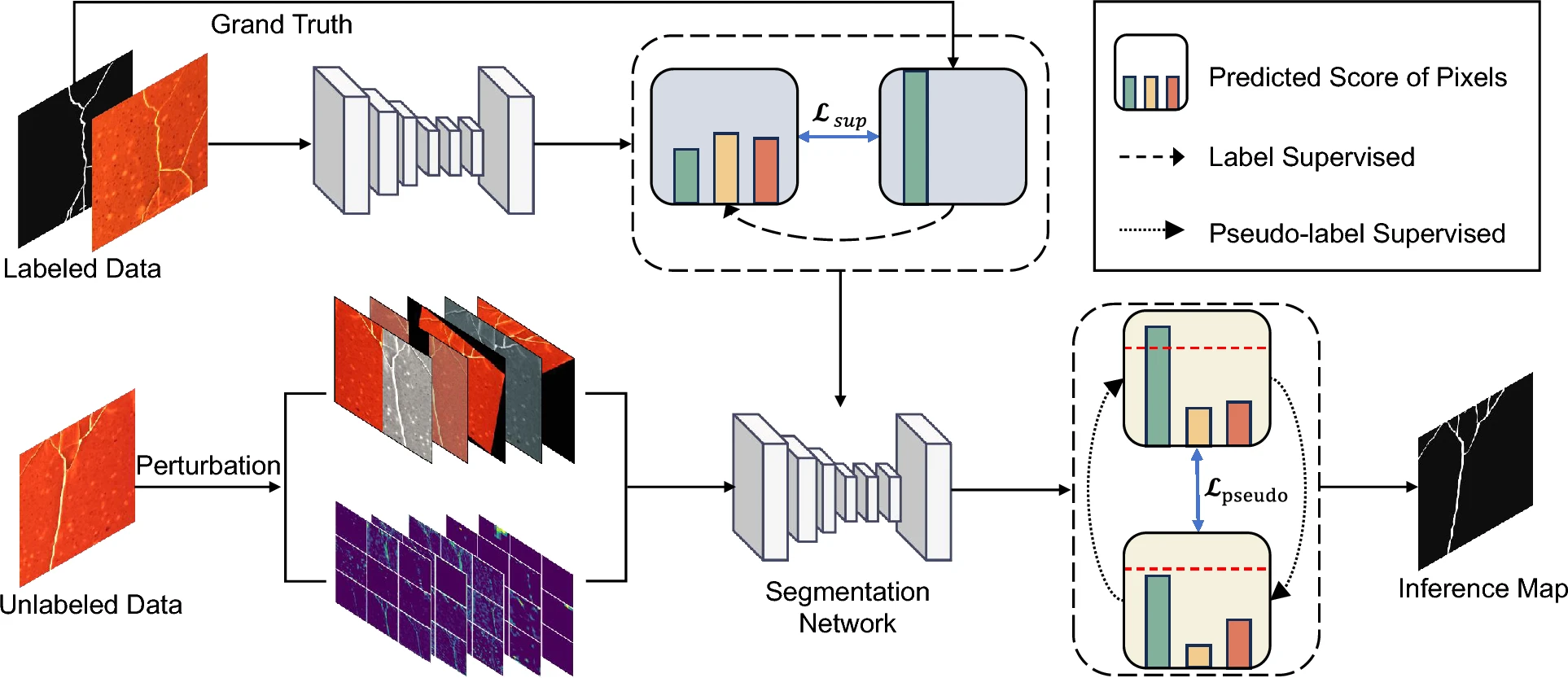}}
\caption{Semi-supervised defect classification model, outlining the modules and the propagation of the labeled and unlabeled samples in the framework \cite{shi2024efficient}.\label{fig4}}
\end{figure}

An additional learning method, semi-supervised learning, incorporates aspects of both SL and USL. The resulting dataset contains both unlabeled data and a small amount of labeled data. Subsequent semi-supervised models can be trained for detection; however, the accuracy of this approach may be lower compared to SL. A semi-supervised classification and detection approach is explored in \cite{10129970}, focusing on semiconductor chip inspection. The minimum ratio of unlabeled to labeled images in the dataset is around 15:1, reaching 120:1 for other classes. The performance of the proposed model shows promise, outperforming baseline supervised models by 2\%–7\% in classification accuracy. Another semi-supervised model applies a cross pseudo supervision framework for the semantic segmentation of industrial defects \cite{shi2024efficient}. Its corresponding framework is visualized in Fig. \ref{fig4}.

\subsection{Scope of Survey}
\begin{figure}[t]
\centerline{\includegraphics[width=3.5in]{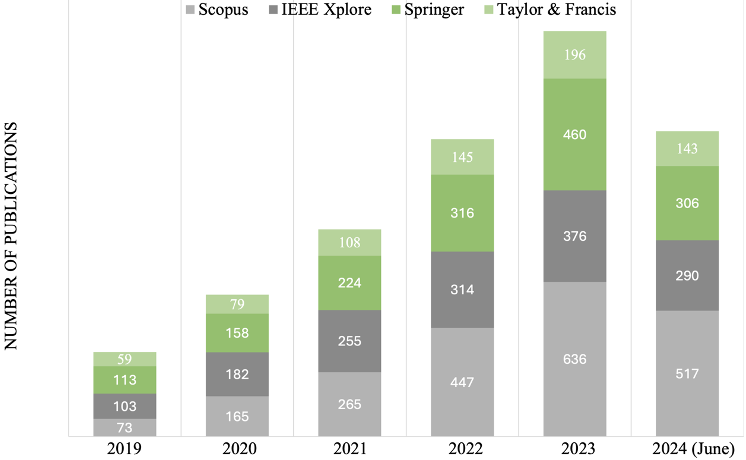}}
\caption{IAD-related publication trends from 2019 to June 2024 and across scientific databases.\label{fig5}}
\end{figure}

To that extent, relevant literature was collected through a keyword-based search on multiple databases. Targeting publications from 2019 onwards, a search was conducted on four scientific databases, including IEEE Xplore, Scopus, Springer, and Taylor \& Francis. These large repositories are pertinent to the review topic and allow for the use of logical operators like "OR" and "AND" while also being able to search in metadata fields including the title, abstract, and keywords. The specific search string was: (“machine learning” OR “deep learning” OR “artificial intelligence” OR “computer vision”) AND (“defect detection” OR “anomaly detection” OR “vision-based defect detection” OR “automated defect detection”) AND (“industrial” OR “manufacturing” OR “industry”). Low-ranked journal papers were excluded from the final selection. Searches revealed that the number of related Industrial Anomaly Detection (IAD) publications grew significantly in recent years, with an increase of nearly 380\% from 2019 to 2023. Fig. \ref{fig5} visually depicts the trend, showing a clear increase in the number of papers up to June 2024. Subsequently, this survey is conducted by presenting the components of an IAD pipeline, including areas related to data acquisition, preprocessing, learning mechanisms, and evaluation. Major areas, including challenges, solutions, applications, and datasets are discussed. Finally, future directions of vision-based IAD are presented, offering researchers insight into the state-of-the-art of industrial inspection. The main contributions of this survey can be summarized as follows:

\begin{itemize}
    \item The collection and organization of relevant vision-based automated inspection publications. Categorizing them in the context of their learning models, applications, and evaluation scores.
    \item Discuss challenges that hinder the adoption of automated inspection in the anomaly detection field. Highlight both scientific and industrial problems, touching upon possible solutions as well.
    \item Showcase the major aspects of an IAD pipeline, from data acquisition to evaluation. Various techniques and methods in each of the four processes are articulated as well.
    \item Explore promising outlooks and future trends in the field of IAD, such as vision language models and explainable AI.
\end{itemize}

\begin{figure*}
\centerline{\includegraphics[width=6in]{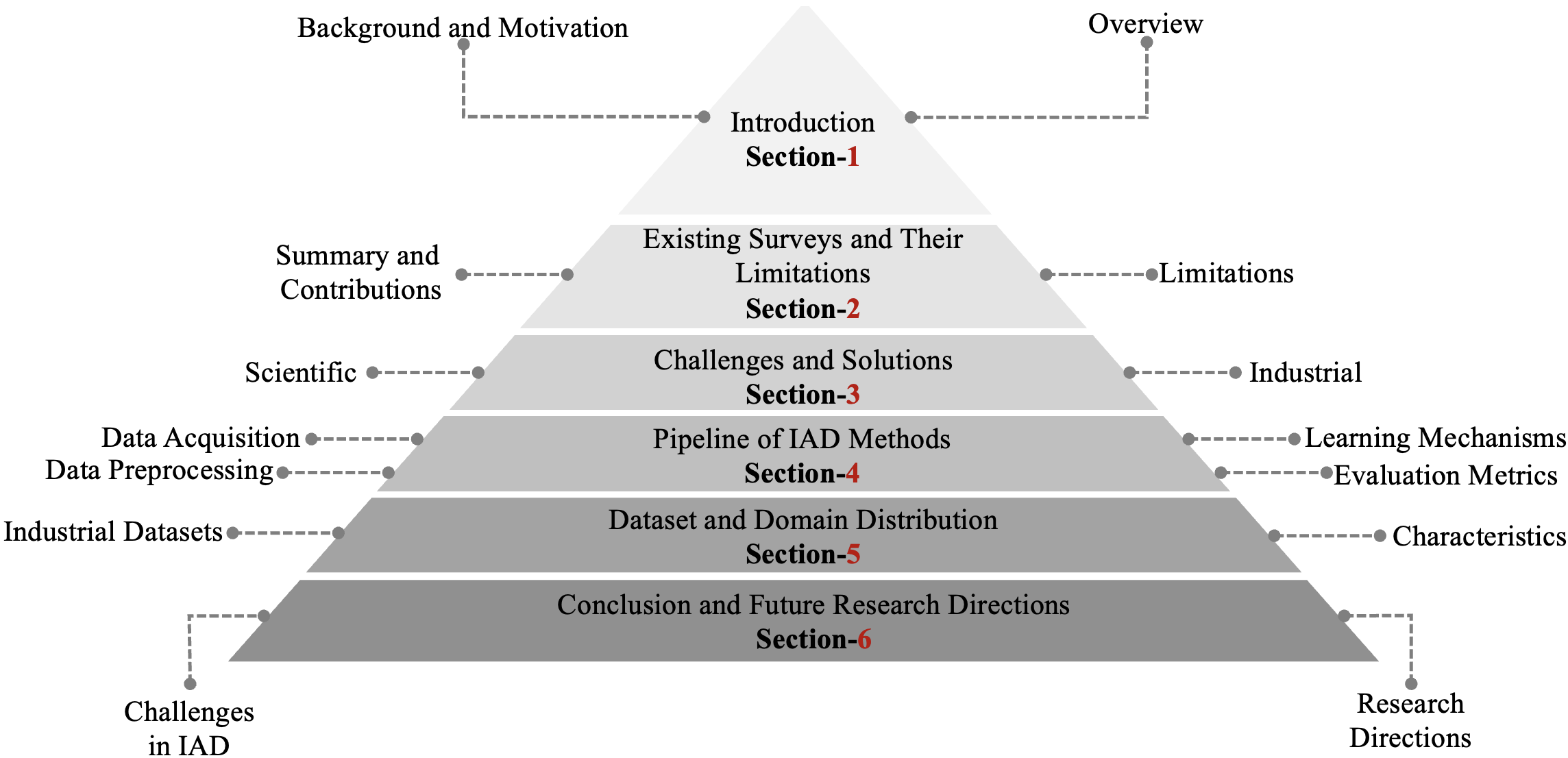}}
\caption{Framework of this survey, showcasing the outlined sections and their respective topics.\label{fig6}}
\end{figure*}

\subsection{Outline}
The rest of the survey paper is organized as mentioned in Fig. \ref{fig6}. Section \ref{EXISTING SURVEYS AND THEIR LIMITATIONS} summarizes the existing surveys in the industrial defect detection field and their limitations. Main IAD challenges and their perspective solutions are presented in Section \ref{CHALLENGES AND SOLUTIONS}. The general pipeline of IAD methods and their applications in terms of data acquisition, data preprocessing, learning mechanisms, and evaluation metrics are introduced in Section \ref{PIPELINE OF IAD METHODS}. Section \ref{DATASET AND DOMAIN DISTRIBUTION} showcases the relevant datasets in the automated inspection domain, with Section \ref{CONCLUSION AND FUTURE RESEARCH DIRECTIONS} mentioning future research directions and concluding the presented survey.

\begin{figure*}
\centerline{\includegraphics[width=\textwidth]{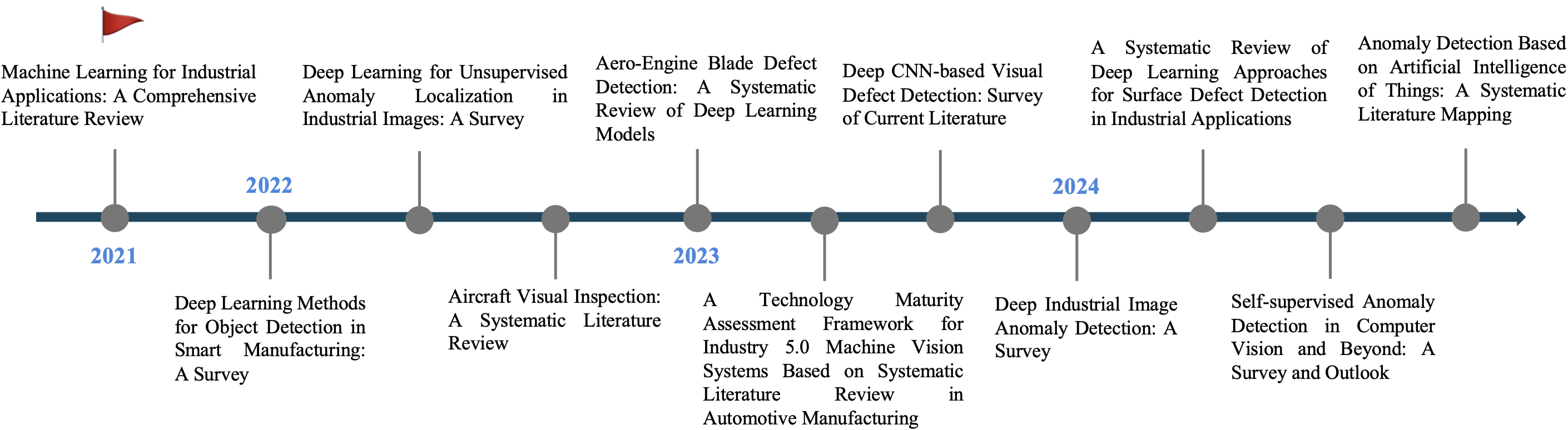}}
\caption{Timeline of published survey articles in the industrial and defect detection fields. The highly cited survey is identified with a red flag.\label{fig7}}
\end{figure*}

\section{Existing Surveys and Their Limitations} \label{EXISTING SURVEYS AND THEIR LIMITATIONS}

Defect detection is a highly studied topic, with numerous articles being published. Survey papers related to IAD, and its various applications, provide a pool of knowledge to the community on the current trends and findings in the literature. Fig. \ref{fig7} showcases some of the recently published surveys in the realm of IAD. Table \ref{table1} summarizes these publications, addressing their contributions and limitations in coverage and scope. 

The diffusion of ML and the potential it offers to solve problems in the operation management field was assessed by \cite{BERTOLINI2021114820}. The review also categorized literature in terms of the applied algorithm (supervised, unsupervised, and reinforcement learning) and application domain, following a keyword search. \cite{AHMAD2022181} Surveyed learning-based object detection methods, including backbone, region selection, and learning strategy. Industrial applications, advantages over traditional methods, and current techniques were discussed. \cite{9849507} provided a comparison of existing anomaly localization methods in industrial images using DL. A review of aircraft inspection techniques was carried out by \cite{YASUDA2022103695}. The authors provided some insight into the incorporation of robotics and computer vision in the visual inspection process. \cite{10138396} Undertook a systematic literature review comparing and contrasting methods, technologies, and performance outcomes of DL models for detecting defects in aero-engine blades. \cite{automanu} Presented a nine-phase systematic review analyzing machine vision works across automotive manufacturing, including 22 different groups of machine vision applications across 47 automotive manufacturing components. Deep CNN-based defect detection models were explored in \cite{JHA2023103911}. Insights into industrial applications, the classification of learning models, and associated challenges were also discussed. \cite{Liu_2024} Reviewed DL-based image anomaly detection techniques, from the perspectives of neural network architectures, levels of supervision, loss functions, metrics, and datasets. \cite{AMERI2024107717} Conducted a recent systematic review of studies in the field of DL-based surface inspection in industrial products. The commonly used datasets for surface defect detection were examined, and a comparative analysis of the performance of DL models was provided. The authors of \cite{HOJJATI2024106106} reviewed the current methodologies in self-supervised anomaly detection. The technical details of the standard methods including their strengths, drawbacks, and comparisonswere discussed. \cite{TRILLES2024101063} presented a systematic literature mapping on the application of anomaly detection techniques in edge computing through microcontroller units.

Among the mentioned publications, \cite{AMERI2024107717} addresses the industrial applications of defect detection. However, it focuses solely on surface-based defects, specifically in DL-based approaches. The previously mentioned surveys also have limitations in scope. Some reviews are broad, lacking discussions specific to industrial applications. Other articles are too specific, investigating single domains. Additionally, a full overview of the IAD pipeline is not presented in similar reviews, due to limited investigations into data acquisition techniques, domain applications, or relevant industrial datasets.

\begin{table}
\caption{Summary of survey papers covering defect detection in industrial fields and their limitations}
\label{table1}
\setlength{\tabcolsep}{3.5pt}
\begin{tabular}{p{34pt}p{35pt}p{25pt}p{58pt}p{58pt}}
\hline
\textbf{Ref/Year} & \textbf{Coverage} & \textbf{DL/ML} & \textbf{Key Contributions} & \textbf{Limitations} \\
\hline
\cite{BERTOLINI2021114820}/2021 & 2000-2019 & \centering \checkmark/\checkmark & Presented the potential of ML and DL in the operation management field. & A bulk of the papers examined predate 2015. Vision-based defect detection is not the primary focus.\\
\hline
\cite{AHMAD2022181}/2022 & \centering - & \centering \checkmark/\ding{55} & Industrial applications and current techniques of object detection methods are discussed. & Addresses object detection in general.\\
\hline
\cite{9849507}/2022 & 2017-2022 & \centering \checkmark/\ding{55} & Presents unsupervised anomaly localization in industrial images. & Only discusses unsupervised ML models.\\
\hline
\cite{YASUDA2022103695}/2022 & 2000-2020 & \centering - & Methods used for the visual inspection of aircraft. & Few publications considered. Limited details are provided about AI defect detection applications.\\
\hline
\cite{10138396}/2023 & 2019-2023 & \centering \checkmark/\ding{55} & Analyzes DL models for detecting defects in aero-engine blades. & Few publications are considered, and they focus solely on DL models.\\
\hline
\cite{automanu}/2023 & 2016-2022 & \centering \ding{55}/\checkmark & Reviews machine vision applications across automotive manufacturing. & Not much focus is placed on defect detection applications, and learning algorithms, along with their evaluation metrics, are not presented.\\
\hline
\cite{JHA2023103911}/2023 & \centering - & \centering \checkmark/\ding{55} & Deep CNN-based defect detection models in industrial applications. & The paper reviews deep CNN-based algorithms.\\
\hline
\cite{Liu_2024}/2024 & \centering - & \centering \checkmark/\ding{55} & Presents architectures related to DL-based IAD techniques. & Does not provide context on broad industrial sectors and their applications of IAD algorithms.\\
\hline
\cite{AMERI2024107717}/2024 & 2020-2023 & \centering \checkmark/\ding{55} & Surveys DL-based surface defect detection in industrial applications. & Focuses on surface detection-based applications only.\\
\hline
\cite{HOJJATI2024106106}/2024 & \centering - & \centering \checkmark/\ding{55} & Reviews current methodologies in self-supervised anomaly detection. & Looks at deep self-supervised learning methods only. Doesn’t specifically address industrial applications.\\
\hline
\cite{TRILLES2024101063}/2024 & \centering 2021-2023 & \centering \checkmark/\checkmark & Mapping the application of anomaly detection techniques in edge computing. &  Analyzes a few papers, with a chunk of them not being image-based models.\\
\hline
\end{tabular}
\end{table}

\section{Challenges and Solutions}
\label{CHALLENGES AND SOLUTIONS}
Vision-based anomaly detection is increasingly being integrated into multiple fields, helping solve inspection and quality control problems. However, automated detection presents unique challenges. Broadly, these complications can be classified as either scientific or industrial. Scientific areas concern the underlying detection model and its functioning, whereas industrial topics relate to the practical application of automated detection in industry. Major problems include real-time detection, complex inspections, annotations, inadequate datasets, data management, and system integration. Addressing these obstacles is not straightforward, as most occur concurrently. Therefore, in an effort to aid researchers in developing more reliable and efficient defect detection systems, prevalent IAD system challenges are summarized.

\subsection{Scientific and Industrial Challenges}
\begin{itemize}
\item \textbf{Real-time inspection} and decision-making are crucial requirements in many industrial sectors. Specific real-time model requirements deal with a model's adaptability to changes, low latency, and seamless processing. This is important because detection systems must keep pace with existing processes. For example, in circuit board manufacturing, an individual component can take less than 10 seconds to pass under an inspection camera and be assembled \cite{electronics13081551}. Furthermore, given the dynamic nature of industrial processes, new anomaly samples may be introduced during real-world operations. Offline learning approaches, therefore, may fail to accurately detect or classify these new instances, as they are trained on fixed datasets. Addressing these problems can involve implementing edge processing. Unlike cloud computing, edge processing brings decision-making data closer to the inspection system, reducing latency and boosting decision-making speeds. Moreover, lightweight defect detection models, such as \cite{10531145}, reduce computational parameters and provide a balance between speed and accuracy. Nonetheless, there are opportunities for continued advancement toward faster inspection and more efficient, lightweight defect detection models.

\item \textbf{Complex inspection} presents the challenges in detecting defects. In real-world settings, defects can vary in size, color, and location, posing significant challenges. In addition, the number and type of defects on a manufactured product are not constant and can occur simultaneously. As a result, defect inspection systems would need to be more complex and applied to a broader set of scenarios. Possible solutions include multi-task learning and the incorporation of numerous data sources \cite{baimachine}. Multi-task learning is used to train models to detect multiple defect classes, improving their understanding of detection and decision-making tasks. Furthermore, the inclusion of more than one input data source, such as normal and thermal images, would grant models fuller inspection capabilities by providing comprehensive and multifaceted information that the other data sources might have missed.

\item \textbf{Small defects and annotation} are persistent challenges in the inspection process. Small defects present a natural challenge as they are not visible to the naked eye, thus they are harder to address when implementing automated inspection. Regardless of their size, the presence of small defects still affects a structure's integrity and makes it susceptible to failure. The difficulty in spotting small defects is due to their relative size to the much larger region of interest. Possible workarounds to this issue include requiring inspection models to have a low false positive rate \cite{9144}, which deals with the number of defective instances being incorrectly classified as non-defective. In addition to small defects, annotated datasets are a constant challenge as well. As presented in Section \ref{DATASET AND DOMAIN DISTRIBUTION}, most datasets employed in IAD have pixel-wise annotations, which are beneficial for precise object detection \cite{9631303}. However, annotation at this fine level of detail across multiple images is costly and time-consuming. Approaches relying on generative models have attempted to circumvent such problems. These solutions involve synthesizing large labeled datasets through generative networks by leveraging a small portion of manually labeled data \cite{li2022bigdatasetgan}.

\item \textbf{Imbalanced and inadequate datasets} affect the prediction and anomaly detection capabilities of models. Obtaining adequate sizes and distributions of industrial datasets is a major challenge for learning-based methods. An imbalance between defective and non-defective samples, with defective samples being harder to acquire, results in a bias towards the more abundant classes. This can also affect performance metric outcomes \cite{10138396}, where high detection scores may be more representative of the larger class and may not correlate to the minority samples. Furthermore, the lack of sufficient training datasets causes lower detection accuracies, due to the model not having much experience in all categories. Remedies for small and imbalanced datasets can involve artificially increasing the number of samples through data augmentation, generative networks \cite{10500332}, and synthetic datasets. Another popular approach, few-shot learning \cite{8954051}, allows models to address data scarcity by training on a very small number of examples. Nonetheless, dataset dependency remains a significant aspect of IAD and is continually being improved.

\item Challenges in \textbf{data quality and management} address the limitations and difficulties surrounding data acquisition. This branch also includes the management of large amounts of data. Adequate solutions require the proper collection, storage, and distribution of data samples. Furthermore, the flow of data is not one-way. The inspection system relies on feedback from evaluations, user inputs, and neighboring systems. The collected data must be of high quality, as low-quality data would undoubtedly hinder inspection. To address these setbacks with quality, suitable sensing and illumination components should be incorporated. Other approaches include the use of adversarial training techniques, which can help improve the robustness of a model, allowing it to perform better under different lighting and imaging conditions. Moreover, the continued progress toward Industry 4.0 will help mitigate these technical and logistical data acquisition challenges.

\item The \textbf{integration} of inspection systems poses an initial challenge for many industries, particularly those with low technological capacities. The growing demand for state-of-the-art automated inspection will introduce new technologies into existing production workflows. A well-designed quality control system is necessary to ensure seamless image acquisition, data flow, ease of maintenance, and overall homogeneity. Moreover, compatibility between newer and older technologies, including underlying protocols and machine-to-machine communication, must also be considered. For instance, IPv6 connectivity should be implemented in industrial settings to enable multiple devices to connect within a network \cite{PHUYAL2020100023}. Nevertheless, a transition from one industrial level to another will require an upfront investment, which will shape future production and labor practices.
\end{itemize}

\section{Pipeline of IAD Methods} 
\label{PIPELINE OF IAD METHODS}
\begin{figure}[t]
\centerline{\includegraphics[width=3.5in]{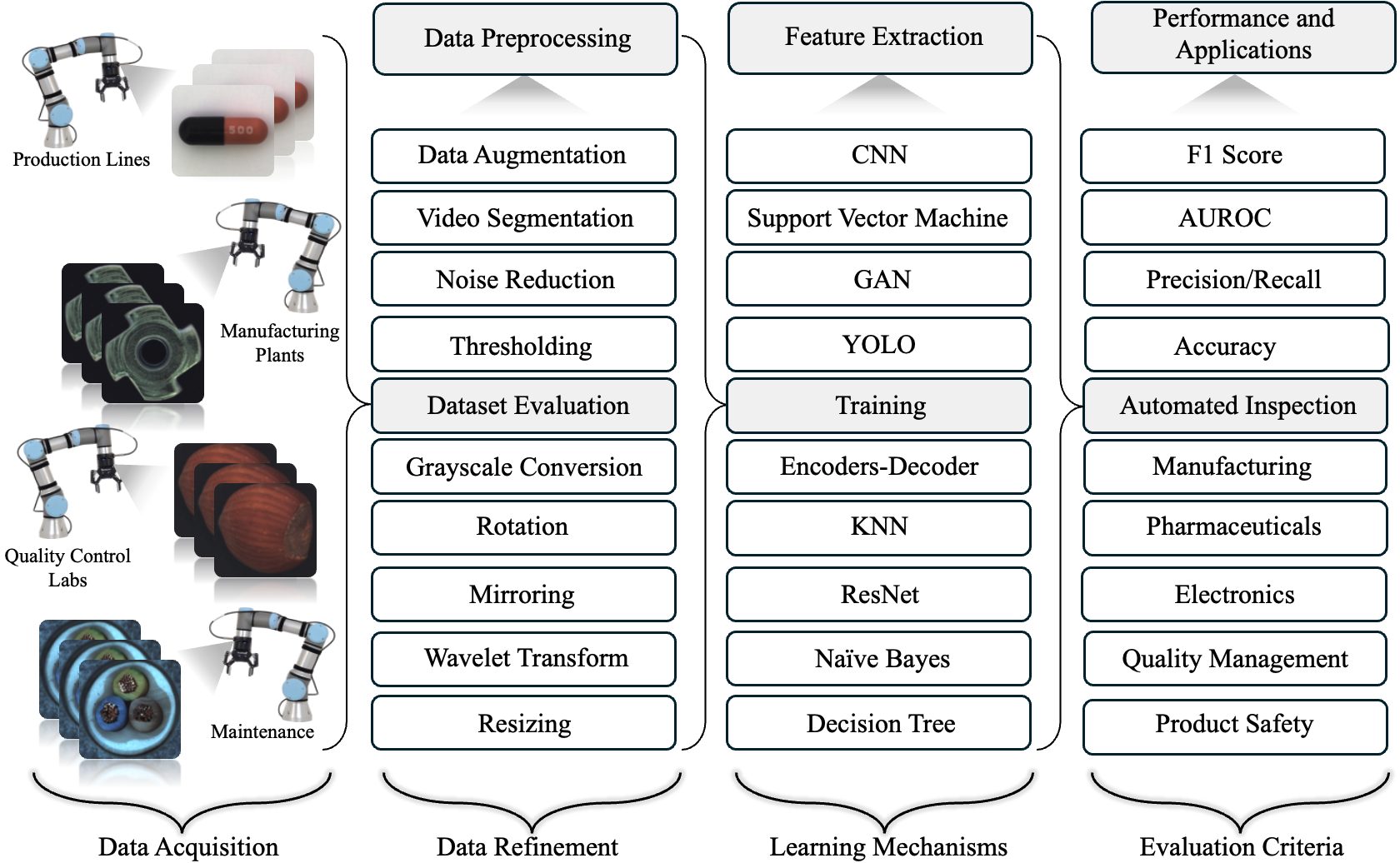}}
\caption{A general pipeline of vision-based IAD. Presents the multiple stages of defect detection and its underlying processes.\label{fig8}}
\end{figure}

This section outlines the fundamental steps involved in IAD processes. The typical pipeline includes data acquisition, preprocessing, feature extraction, and evaluation. Fig. \ref{fig8} visually represents this overall workflow. Data is the first aspect considered in any inspection scenario due to its significance. During acquisition, images are collected through methods such as line-scan imaging or thermography and they contain vital information about the object of interest. To make this raw data useful and meaningful as model input, preprocessing techniques are carried out. Image quality is also improved during preprocessing by enhancing specific features and reducing distortions or noise. Next, features are extracted from the processed images using learning models to capture and highlight critical attributes related to defects, objects, and the environment. Additionally, depending on the input and desired output, various learning methods can be employed at this stage. Lastly, the evaluation phase assesses the model’s output, predictive accuracy, and reliability, often using metrics such as accuracy, precision, and AUROC. This overall pipeline and its components are essential in IAD applications. Table \ref{table2} overviews additional published studies in industrial automated inspection. The models, applications, and levels of supervision are highlighted.

\begin{table*}[t]
\centering
\caption{Publications on vision-based learning models. SL: supervised learning. USL: unsupervised learning. SSL: semi-supervised learning.}
\label{table2}
\setlength{\tabcolsep}{3.5pt}
\begin{tabular}{p{20pt}p{40pt}p{35pt}p{380pt}}
\hline
\centering \textbf{Ref} & \centering \textbf{Model} & \textbf{Learning Strategy} & \textbf{Contributions}\\
\hline
\centering \cite{10382509} & \centering ResNet152 \& YOLOv5s & \centering SL & $\bullet$ Inspection of fatigue cracks in additively manufactured Ti-6Al-4V parts. \newline $\bullet$ Geometrical characteristics are used in decision making. \newline $\bullet$ Achieved 82.3\% detection of initial defect sites.\\
\hline
\centering \cite{YANG2022108338} & \centering NDD-Net & \centering SL & $\bullet$ Detection of defects in weld and rail images. \newline $\bullet$ The segmentation framework included an encoder-decoder base, an attention fusion block, and a residual dense connection convolution block. \newline $\bullet$ Scores showed better performance compared to other models, with a Dice score of 0.841.\\
\hline
\centering \cite{9449912} & \centering CycleGAN \& U-Net & \centering USL & $\bullet$ Two-stage framework for the generation and pixel-wise segmentation of textured surface defects. \newline $\bullet$ Computationally efficient and required a few real defect samples, without any manual annotations. \newline $\bullet$ A precision of 93.35\% made it ideal for real-time inspection in manufacturing settings.\\
\hline
\centering \cite{ruff2019deep} & \centering Deep SAD & \centering SSL & $\bullet$ General anomaly detection on numerous datasets.  \newline $\bullet$ Incorporates both normal and anomalous labeled samples for training.  \newline $\bullet$ The model achieves higher AUC averages than hybrid and deep techniques.\\
\hline
\centering \cite{badmos2020image} & \centering VGG19 & \centering SL & $\bullet$ Inspection of lithium-ion battery electrodes through a pretrained convolution model. \newline $\bullet$ Classification tests demonstrated an F1-score of 0.99.\\
\hline
\centering \cite{9537583} & \centering Multi-task deep one-class CNN & \centering USL & $\bullet$ The defect classification and detection model did not use any defective or annotated samples. \newline $\bullet$ Tests conducted on an aerospace weld dataset had an accuracy of 87.8\%\\
\hline
\centering \cite{10555503} & \centering YOLOv5s & \centering SL & $\bullet$ Inspection of surface PCB defects, including six types of anomalies. \newline $\bullet$ A CoTNet transformer module is used for feature extraction, while a global attention mechanism enhances the model's feature learning. \newline $\bullet$ The algorithm achieves a mean average precision of 98.5\%.\\
\hline
\centering \cite{s24010232} & \centering ResNet & \centering SL & $\bullet$ Defect detection of ceramic pieces in an industrial setting. \newline $\bullet$ Using binary classification, the model included image acquisition and data storage within its framework. \newline $\bullet$ An accuracy of 98\% was attained when the system was placed in a stoneware factory.\\
\hline
\centering \cite{9007362} & \centering OC-SVM & \centering USL & $\bullet$ ML-based model for the detection of faults in urban sewer pipelines. \newline $\bullet$ Video data were converted into feature vectors and used to evaluate a isolation forest model. \newline $\bullet$ An overall accuracy of 90.2\% was reported.\\
\hline
\centering \cite{9121251} & \centering ResNet & \centering SSL & $\bullet$ Proposed industrial surface inspection method. \newline $\bullet$ Incorporates advanced data augmentation using MixMatch and a novel loss function. \newline $\bullet$ The outlined method achieved good performance in comparison with benchmarks.\\
\hline
\centering \cite{LI2023102470} & \centering YOLOv4 & \centering SL & $\bullet$ Three aspects of the model include the one time classification and regression of data, the residual structure, and a feature pyramid network. \newline $\bullet$ Tested on a dataset of manufactured parts, the model had a mean average precision of 94.5\%.\\
\hline
\end{tabular}
\end{table*}

\subsection{Data Acquisition}
An inspection system's accuracy heavily relies on the quality and quantity of data. As mentioned, traditional human visual perception is limiting in its applications in defect detection. For instance, an operator’s vision can become fatigued after prolonged periods. Inspectors also require training and experience and can only view a portion of the electromagnetic spectrum. Thus, reliance on human vision alone is not adequate. As mentioned before, vision-based systems can enhance established quality inspection. Today, high-resolution, cameras are offered at reasonable prices and can be integrated into numerous smart manufacturing operations \cite{BABIC2021262}. Machine vision technology takes in a range of wavelengths which allows computers to visually interpret their environment \cite{arviso}. Paired with sophisticated optical sensors and image processing algorithms, inspection models can perform operations beyond the capabilities of human operators. As computer technology and artificial intelligence have progressed, machine vision has started to be used in production processes and has become an effective inspection method \cite{ainspec}. This technology enables automation and enhances detection while reducing manpower requirements in industrial processes \cite{abdulrahman2022ai}.

\begin{figure}[b]
\centerline{\includegraphics[width=3.5in]{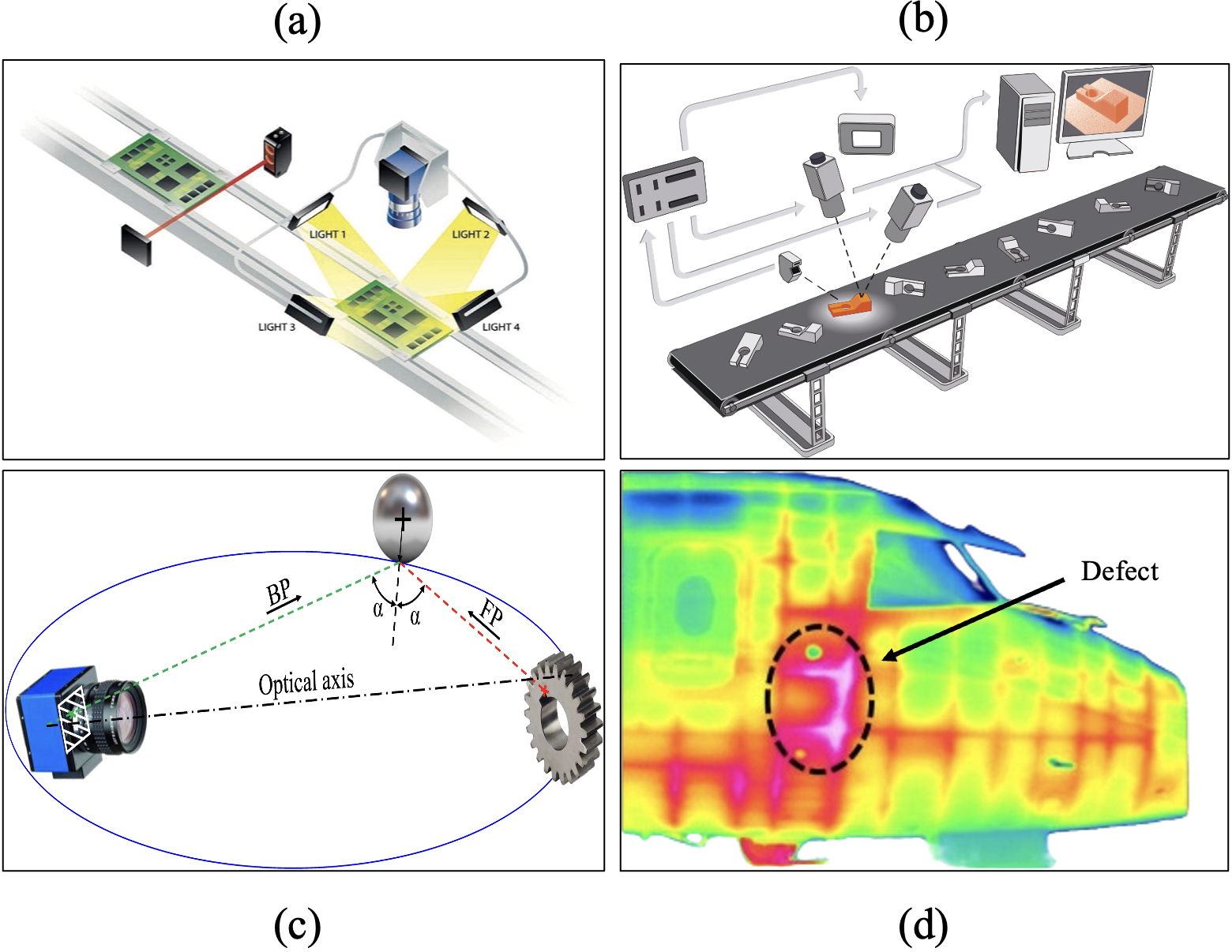}}
\caption{Imaging configurations and techniques utilized for data acquisition. (a) Line-scan imaging \cite{noauthor_need_2018}. (b) Multi-view imaging \cite{noauthor_3d}. (c) Omnidirectional vision \cite{s18020408}. (d) Infrared thermography \cite{BAGAVATHIAPPAN201335}.\label{fig9}}
\end{figure}

In practical industrial processes, various imaging modalities are employed, with the choice of method depending on the specific application and environment. In most manufacturing settings, large volumes of standardized products move through assembly lines during production. In these cases, imaging is typically performed using line-scan cameras mounted directly above a conveyor. For example, leather inspection is carried out as it passes under a line-scan camera \cite{electronics11152383}. However, not all industrial fields can or do acquire data in this way. A constant challenge for visual inspection is the ability to capture all key information from a single viewpoint. For objects with complex anatomies, multiple images are required to detect any present defects \cite{mulview}. In the inspection of a vehicle’s exterior, researchers employed a multi-view detection system to address this challenge \cite{carview}. The model captured various viewpoints of the car and merged them into a single 3D image. This allowed a single-view learning model to be implemented without the need to alter the existing model or data. When multiple cameras are not feasible and the inspection system size must be limited, omnidirectional vision can be used. This method enhances the field of view with a wide-angle lens to capture surroundings in all directions \cite{10.1007/978-1-4471-1580-9_18}. It is commonly used for detecting defects inside pipes, as demonstrated in \cite{ZHANG2024108003}, where a YOLOv8 model with catadioptric sensors detected and classified pipeline images. Furthermore, due to panoramic image distortion, an unwrapping technique was applied before processing and the model achieved 80\% detection accuracy. An additional imaging technique is infrared thermography. Infrared cameras visualize the temperature distribution of an object’s surface \cite{s23094444}. Also known as thermograms, these temperature distributions can provide insight into the magnitude and location of defects in internal structures \cite{s23218780}. Infrared imaging is commonly used in the photovoltaic module \cite{pvarticle} and electronics industries \cite{WANG2023112307}. Examples of each of the four imaging techniques are shown in Fig. \ref{fig9}.

\subsection{Data Preprocessing}
Prior to model training, collected data must undergo preprocessing, which, in most cases, encapsulates changing the size, color, and orientation of captured data. Furthermore, during this step, video data is sequenced into frames, where it can be processed and have its quality enhanced. Proper preprocessing can lead to more fruitful detection outcomes. Additionally, factors such as a camera’s view, the transformation of optical signals, and electronic components can introduce noise in data. The presence of noise affects data quality and negatively impacts inspection. Accordingly, the overall job of data preprocessing is to remove noise and highlight important sections of data. Data augmentation, on the other hand, helps with the problem of insufficient training data that is plaguing many industrial fields. This is significant since learning-based models require large amounts of data. By altering raw captured images, different images can be created and added to the training dataset, thus increasing the data volume. Various data preprocessing and augmentation methods are employed in different industries, including grayscale conversion, sharpening, rotation and mirroring, resizing, thresholding, and wavelet transform.

Grayscale conversion removes all colors from an image, leaving only shades of gray. This is carried out by transforming the 24-bit RGB pixel values to 8-bit grayscale values instead \cite{5445596}. Firstly, this significantly reduces computational requirements, as converting an image to grayscale reduces it to its lowest level of pixels. Secondly, the contrast of the image can be enhanced, possibly allowing for easier detection of defects. Sharpening an image is a digital photo effect used to enhance clarity. This process makes the edges of the image more defined and crisp \cite{10311554}. Sharpening is especially important if the captured data is blurry, as it helps to distinguish the object under inspection from the background, thereby avoiding errors. Image rotation is a common technique in image processing that rotates images by some angle in any direction. Parallel to this, mirroring may also be used to generate more data samples. Mirroring reflects an image by changing pixel locations about a plane, either horizontally or vertically. These techniques also help diversify the dataset by using the existing images. In cases where model training times are an issue, data resizing is often employed. Decreasing the proportions of large images can significantly accelerate model training while keeping performance relatively the same \cite{10466741}. Resizing also decreases memory and computational needs. Thresholding is a popular technique applied in machine vision defect detection \cite{NG20061644}. It is a straightforward and powerful method for dividing an image into a foreground and background. Given that grayscale representations of defected regions are typically brighter or darker than those of the environment, thresholding techniques can effectively isolate defects from non-defective surroundings. It accomplishes this by transforming entities in a grayscale image into a binary image. This allows inspection models to differentiate between defects and normal regions more easily, with limited noise and background pixels. Another widely used image preprocessing algorithm is wavelet transform. This technique precedes thresholding and segregates the signal (image) into different frequency scales, allowing for the analysis of each scale to determine time-frequency attributes. It is effective for feature extraction, noise reduction, and image improvement \cite{UESUGI2023103442}.

\subsection{Learning Mechanisms}
The progress in defect detection has been driven by the rise of learning-based methods. Consequently, several IAD techniques and applications have emerged. This section discusses some of the various learning-based models and algorithms utilized in the IAD and vision-based detection fields. Both ML and DL based algorithms are included in this discussion. These models are employed in the analysis of images to accurately identify defects. Additionally, they are constantly being improved and altered for use in differing industrial applications. This section also provides a summary of published literary works on IAD, highlighting techniques and implementations.

ML algorithms, a subset of artificial intelligence, are systems that work on a task while learning and teaching themselves new information. One such algorithm, SVM, is widely used for classification tasks across industries. The generalization ability of SVM has received significant attention from pattern identification researchers \cite{CERVANTES2020189}. It operates by constructing an optimal decision boundary that separates classes within a dimensional space. Thus, new data can be categorized into one of the respective groups created by the SVM. Several hyperplanes can be created for multiple classes, which are constructed by support vectors, explaining the SVM name. The K-Nearest Neighbor (KNN) algorithm is a supervised ML classifier that makes predictions based on the proximity of individual data points to each other. KNN works on a simple principle and organizes data into groups and classifies any new input data based on its likeness to previous data \cite{9065747}. Data is classified based on its closeness to the neighboring groups, calculated by some distance metric, which is most commonly Euclidean distance. KNN falls under “memory-based learning” because it stores all its training data in memory. This hinders the efficiency of KNN on larger datasets, due to longer processing times and computation costs. Naïve Bayes (NB), known for its fast and independent classification of data, aids models that require the ability to make speedy predictions. Based on Bayes’ theorem, NB assumes that features are independent of the presence of other features. This gives it the advantage of being able to train on smaller datasets and deal with multiple classes. However, non-correlation between features is not observed in real-world applications. If NB is implemented while feature dependency is ignored, performance can be negatively affected \cite{jourNB}. Decision tree is another important supervised ML method employed in classification and prediction \cite{9918857}. Decision trees resemble a tree-like or flowchart structure, beginning as a starting node with a specific data question that branches out to a possibility of answers. These branches continue to decision nodes that pose additional questions, resulting in further branches. This process continues until the data is at an end node location, representing a prediction or output. Decision trees can give insight into the decision-making process, allowing researchers to analyze all possible outcomes. However, due to their structure, end results can be completely different if the starting data is altered even slightly. The implementation of multiple decision trees is known as a random forest algorithm. It is another classification technique and works by merging the output results from many decision trees using the bagging method. Bagging, a type of ensemble learning, combines several estimates from numerous models and incorporates them into a single model with better stability and accuracy. However, the generalization error of the random forest still depends on the individual strength of the decision trees and their correlation with each other \cite{breiman_random_2001}.

While ML interprets information from data through simpler means, DL relies on an Artificial Neural Network (ANN) for decision-making. These neural networks analyze data through a logical structure, imitating the biological process of the human brain. The neural networks are composed of layers of nodes that contain input, output, and hidden layers. The specific number of layers and nodes depends on the complexity of the network, with nodes being interconnected from one layer to another. As data passes through the ANN, the neural network gradually learns more information about the data until it outputs the results. A very popular DL algorithm, Convolutional Neural Network (CNN), is commonly used for vision-based tasks \cite{defCNN}. They automatically identify and extract important features from images by applying a series of convolutional filters. The primary blocks of a CNN include convolution layers, pooling layers, and fully connected layers. Convolution layers identify key patterns in images, also known as feature maps. The pooling layer then reduces the dimensions of the input, ensuring only relevant information is kept. A Generative Adversarial Network (GAN) is an unsupervised DL algorithm that has found uses in machine vision defect detection tasks because of its strong generative capabilities \cite{articlegancite}. This architecture is used for generative tasks such as image synthesis by automatically detecting features from input data and producing new samples based on those features. GAN consists of two networks - a generator network that creates new data and a discriminator network that distinguishes generated data from real data. The algorithm generates region proposals in an image and uses CNN to classify each proposal. Another DL-based algorithm is You Only Look Once (YOLO), which grants machines the human ability to interpret an image at a glance \cite{7780460}. This algorithm uses an end-to-end neural network that makes predictions of bounding boxes and class probabilities all at once. YOLO has a simple and compact model structure allowing for fast processing speeds \cite{JIANG20221066}. This enables real-time inspection capabilities for more practical applications of automated inspection \cite{10539065}. The encoder component works by taking an input image and capturing essential features, which are reconstructed by the decoder into a new image or other forms of data. Examples of this architecture show improved feature extraction and contextual understanding \cite{9875035}. The Residual Neural Network (ResNet), first introduced in 2015 \cite{7780459}, is a DL-based model adopted widely in image recognition tasks. A ResNet can overcome the problem of degradation. This problem occurs as a neural network becomes deeper (i.e., more complex). With more layers added, the performance of the network becomes affected. ResNet can overcome this limitation while extracting more features from data. The ResNet architecture is made of residual blocks that allow a neural network to have many more layers without this drawback. This is accomplished by skip connections or shortcuts which, in essence, permit for the bypassing of one or more layers of the network.

Other IAD methods exist that are more specifically considered for industrial settings. These methods include few-shot learning, teacher-student models, noisy learning, and one-class classification. For cases where labeled training data is scarce, few-shot learning can be employed \cite{Parnami2}. This learning technique is a subset of a broader approach known as “n-shot learning”. Since data collection and labeling are problems in many real-world applications, few-shot learning can reduce the need for annotation and compilation of industrial product images when training models. Furthermore, few-shot learning can be used to train pre-trained models on new classes of data, saving time and computational resources. A few-shot framework is proposed by \cite{9623623} for steel surface defect detection. A pre-trained ResNet-50 was implemented as the backbone of the framework and trained on a public steel surface defect detection dataset. The experimental results showed that the proposed framework was able to perform on par with models trained using abundant data. Furthermore, other parallel learning methods are explored in the literature. Most notably, zero-shot learning, an extension of n-shot learning, is trained to recognize and classify data without having seen it before. This overcomes the need for exhaustive training data compilation and labeling. WinCLIP \cite{10204096}, which is based on the vision-language model CLIP \cite{radford}, implements zero-shot image anomaly detection. Utilizing CLIP’s image and text encoders for multi-scale feature extraction, WinCLIP compares the extracted vision and text features and checks the similarity between them to identify any present defects. WinCLIP outperforms prior few-shot works on IAD datasets, with it improving upon PatchCore \cite{9879738} by 9.7\%. Another learning method, teacher-student, is based on knowledge distillation, which involves transferring information from a larger model to a smaller one. In this case, the larger teacher model is trained on a substantial dataset and distills its knowledge to the smaller student model. While the teacher is trained on diverse instances of data, which is computationally and memory demanding, the student can focus on specific information, making it more suitable for deployment \cite{Hinton2015DistillingTK}. The goal is to then allow the student model to achieve comparable results to the cumbersome teacher model \cite{hu2023teacher}. A novel teacher-student anomaly detection model is explored by \cite{9157778}. The authors propose an unsupervised anomaly detection framework using industrial image datasets like MVTec, where a student model ensemble is trained on a defect-free dataset to imitate a pre-trained teacher model. A major complication in many anomaly detection datasets is noise. Noise can cause uncertainties in learning models by making it harder to identify patterns and boundaries in objects. In real-world IAD applications, noise can cause many of these issues due to the complexity of the products and inspection processes involved. Remedies to polluted data include efforts in denoising data at the patch level. A proposed novel anomaly detection algorithm, SoftPatch \cite{jiang2022softpatch}, looks at patch-level denoising in image samples. Tests on two IAD datasets at varying noise settings showed the SoftPatch algorithm achieving optimal results compared to other state-of-the-art anomaly detection methods. Nevertheless, poorer performance caused by SoftPatch not considering feature distribution and its higher sensitivity to noise leaves room for future research and development in noisy anomaly detection methods. Table \ref{table3} collects and presents additional IAD-related publications and models, focusing on their specific deployment.

\begin{table*}[t!]
\centering
\caption{IAD model publications, highlighting their application in the industrial field as well as their respective platforms.}
\label{table3}
\setlength{\tabcolsep}{3.5pt}
\begin{tabular}{p{15pt}p{190pt}p{190pt}p{75pt}}
\hline
\centering \textbf{Ref} & \centering \textbf{Description} & \centering \textbf{Contribution} & \textbf{Platform}\\
\hline
\centering \cite{9306873} & Proposed self-adaptive method for real-time PCB defect detection. The model is continuously retrained on edge data and redeployed in the inspection line, improving detection accuracy and reducing false alarms. & Shortens inspection and repair times in real-world production lines, boosting PCB factory efficiency by 33\%. Eliminates manual data labeling and enables operators to repair solder points based on model-generated highlights. & Automated Optical Inspection (AOI) edge and a server for model training\\
\hline
\centering \cite{10472782} & A proposed inspection system for manufactured hydrogen storage tanks using a CNN and an unsupervised clustering model. Incorporates an augmented hybrid learning method, training only on non-defective samples. & Achieves 90\% recall, outperforming benchmarks while maintaining low false positives. Tested on a real-world hydrogen tank production line, the model has proven more efficient than human inspectors in detecting internal surface defects. & Edge computer server\\
\hline
\centering \cite{9530369} & 
An in-situ inspection system for additively manufactured metals is applied in a real-world industrial setting. The GAN-based approach addresses data scarcity and a microservice architecture enables scalability. & The system identifies five defect categories in real time, incorporating layer-wise monitoring of the metal process. Its plug-and-play framework simplifies the integration of Industry 4.0 devices. & Platform as a Service (PaaS)\\
\hline
\centering \cite{Tabernik2019JIM} & A DL framework detects and segments surface defects on electrical commutators. The two-stage decision and segmentation network performs pixel-wise detection and classification in a binary manner. & The architecture is adaptable to new domains without modification. Datasets of only 33 samples are used, reducing computational and data collection costs. It is comparable to other DL methods, showing only one miss-classification. & TensorFlow\\
\hline
\centering \cite{bugatti2022towards} & In-situ detection system incorporates a high-speed camera for capturing videos during metal manufacturing. The data extraction algorithm features thresholding and region isolation. & Identifies regions of interest and significantly reduces data dimensions. Experiments demonstrated the framework's real-time detection capabilities, spotting defects in approximately 35 ms. & HPC\\
\hline
\centering \cite{MUJEEB2019100933} & A DL inspection system employs an autoencoder for image feature extraction. Data augmentation, such as flipping and rotation, were applied to create a training dataset of over 400 samples from a single PCB reference image. & The lightweight model serves as a low-cost, flexible solution and is trained exclusively on non-defective samples. Another parameter, block size, uniformly divides large images into smaller sections, enhancing inspection suitability. & AOI edge system and a server for model traning\\
\hline
\centering \cite{XIA2020845} & Inspection of welded parts using a residual network on five classes. The setup uses a fixed HDR camera to capture images of moving steel pieces. The training set for each class comprises 1,000 images, including augmented samples. & The 18-layer convolutional network has a center loss module connected to the last pooling layer for optimized training. Skip connections expedite model training, whereas guided gradient-class activation interprets the model's decisions. The system achieved 98\% accuracy. & Google Colab\\
\hline
\centering \cite{davtalab2022automated} & A CNN model is introduced for the automated detection of layer defects in printing concrete. A total of one million labeled images are used in the training and testing datasets. The model can differentiate between pixels on various concrete wall geometries. & The model consists of an autoencoder and a pixel-wise segmentation algorithm. To handle the large dataset, an edge detection algorithm was included. An F1-score exceeding 90\% was obtained when testing on new images. & TensorFlow\\
\hline
\centering \cite{zhu2020vision} & A transfer learning and CNN-based approach is proposed in the inspection of bridges. Images are captured and input into a built-in user interface to provide real-time results. & The transfer learning model requires much less computational power and training time. Images of varying sizes and qualities can be accepted by the system. Testing on a separate dataset resulted in an accuracy of 97.8\%. & User interface edge and TensorFlow\\
\hline
\centering \cite{MENDOZABERNAL2024123210} & This study presents an approach for detecting and classifying defects in agricultural crops, using multiple CNN architectures chosen for their excellent classification performance. & Dataset challenges were addressed through augmentation, and Gaussian noise was applied to improve model generalization. Testing on two datasets resulted in accuracies of 98\% and 95.3\%, surpassing the state-of-the-art. & DEEP-Hybrid-DataCloud\\
\hline
\end{tabular}
\end{table*}

\subsection{Evaluation Metrics}
\begin{figure}[b]
\centerline{\includegraphics[width=3.5in]{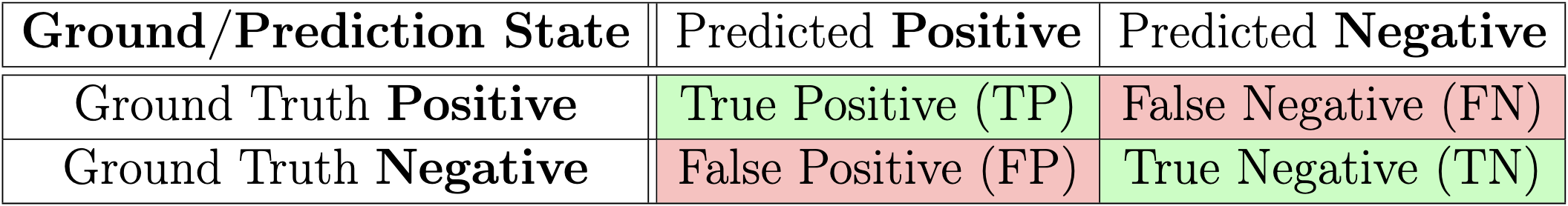}}
\caption{Confusion matrix used to define the performance of a classification algorithm.\label{fig10}}
\end{figure}

The evaluation of machine vision models and their outputs is essential. The essence of any model lies in its reliability and validity, as they play a crucial role in confirming the credibility and accuracy of the results. Most evaluation metrics are based on parameters stemming from the comparison of the prediction results and ground truth. If a model detects a defect during inspection, and that defect is present in the ground truth, it is considered a True Positive (TP). On the other hand, if the model identifies a defect that does not truly exist, it is categorized as a False Positive (FP). Conversely, if a defect present in an image fails to be identified, it is labeled as a False Negative (FN). Lastly, if the model correctly identifies the absence of a defect, and indeed no defect is shown, it is known as a True Negative (TN). These terms can be broken down into a matrix, referred to as a confusion matrix, which is presented in Fig. \ref{fig10}. Based on this, the commonly practiced evaluation metrics across vision-based defect detection include precision, accuracy, recall, specificity, F1 score, AUROC, and Intersection Over Union (IOU). Formulas for these metrics are presented in Table \ref{table4}, whereas Table \ref{table5} contains the reported findings of published industrial inspection models and their respective inspection applications.

Precision depicts the ratio of true positive predictions in relation to the overall number of predictions made. Thus, it assesses the model’s accuracy in correct positive predictions. This gives insight into the model’s ability to differentiate between important details (defects) and irrelevant aspects. Accuracy gives a broader evaluation by representing the correctly classified samples. It gives a simple percentage of the correctly classified data in a dataset. Accuracy can serve as an initial evaluation of a model’s performance, but it can be misleading. For industrial datasets with class imbalances, accuracy might suggest inspection performance that is not true for all defect classes. Recall focuses on quantifying how accurate a model is in identification. This is crucial for identifying defects, where a high recall indicates a model’s ability to find objects of interest. Specificity, on the other hand, measures how many true negative samples are correctly classified by the model. High specificity implies that the model is correctly predicting most of the negative results (i.e., non-defective samples). The F1 score combines both precision and recall through their harmonic mean and ranges between 0 and 1. As precision and recall scores diverge from each other, the harmonic mean decreases, reflecting a poorer overall score. A high F1 score indicates a better classifier model. The Receiver Operating Characteristic (ROC) is a curve that represents the recall (true positive rate) and the value of 1 minus the specificity (false positive rate) plotted at different criteria. This gives a graphical representation of a model’s ability to differentiate between positive (defective) and negative (non-defective) classes over various thresholds. The area under the ROC curve, AUROC, is a single metric that quantifies overall classification performance. An AUROC of 0.5 reflects a model that classifies results randomly, whereas a value of 1 indicates a perfect classifier. The concept of IOU is applied to determine if an area of interest (e.g., defect) was localized correctly. The metric takes into account two bounding boxes: one representing the ground truth and the other representing the predicted bounding box. A resulting IOU of 1 reflects a complete overlap of the predicted and ground truth (i.e., area of interest is correctly contained).

\begin{table}[hb]
\centering
\caption{Common vision-based evaluation metrics. TP: True Positive. TN: True Negative. FP: False Positive. FN: False Negative. TPR: True Positive Rate. FPR: False Positive Rate.}
\label{table4}
\setlength{\tabcolsep}{3.5pt}
\begin{tabular}{p{35pt}p{150pt}}
\hline
\centering \textbf{Metric} & \textbf{Formula}\\
\hline
\centering Precision & 
${TP}/{(TP+FP)}$\\
\hline
\centering Accuracy & 
${(TP+TN)}/{(TP+TN+FP+FN)}$\\
\hline
\centering Recall & 
${TP}/{(TP+FN)}$\\
\hline
\centering Specificity & 
${TN}/{(TN+FP)}$\\
\hline
\centering F1 Score & 
${TP}/{(TP+\frac{1}{2}(FP+FN))}$\\
\hline
\centering AUROC & 
$\int_{0}^{1}{(TPR)d(FPR)}$\\
\hline
\centering IOU & 
{Area of Overlap}/{Area of Union}\\
\hline
\end{tabular}
\end{table}

\begin{table}[t]
\centering
\caption{Reported evaluation scores of automated inspection models used in IAD-related applications.}
\label{table5}
\setlength{\tabcolsep}{3.5pt}
\begin{tabular}{p{20pt}p{40pt}p{25pt}p{30pt}p{30pt}p{25pt}p{20pt}}
\hline
\centering \textbf{Ref} & \centering \textbf{Inspection Application} & \centering \textbf{F1-Score (\%)} & \centering \textbf{Accuracy (\%)} & \centering \textbf{Precision (\%)} & \centering \textbf{Recall (\%)}& \textbf{IOU (\%)}\\
\hline
\centering \cite{amieghemen2023deep} & \centering Pavement & \centering 98.03 & & \centering 98.03 & \centering 96.07 & \hspace{0.1cm} 96\\
\hline
\centering \cite{huang2022deep} & \centering Subway tunnel lining & \centering 68.68 & \centering 81.94 & & & 52.72\\
\hline
\centering \cite{s21154968} & \centering PCB & \centering 98 & \centering 98 & \centering 98.3 & & \\
\hline
\centering \cite{chen2020robust} & \centering Industrial surfaces & \centering 74.4 & & \centering 77.69 & \centering 71.5 & 70.56\\
\hline
\centering \cite{iqbal2024multi} & \centering Aero-engine & & & \centering 91.4 & \centering 77.6 & \\
\hline
\centering \cite{9736943} & \centering Drum-shaped rollers & \centering 98.8 & & \centering 99.2 & \centering 98.9 & \\
\hline
\centering \cite{zhang2023automated} & \centering Bridge & \centering 92 & & \centering 93.96 & \centering 90.12 & \\
\hline
\centering \cite{10416966} & \centering Paper & \centering 93 & & \centering 90.9 & \centering 85.9 & \\
\hline
\centering \cite{liong2020leather} & \centering Leather & \centering 94.76 & \centering 94.67 & & & \\
\hline
\centering \cite{XU2024124244} & \centering Zippers & \centering 98.18 & & \centering 97.96 & \centering 98.4 & \\
\hline
\end{tabular}
\end{table}

\section{Dataset and Domain Distribution} 
\label{DATASET AND DOMAIN DISTRIBUTION}
\begin{figure}[b!]
\centerline{\includegraphics[width=3.5in]{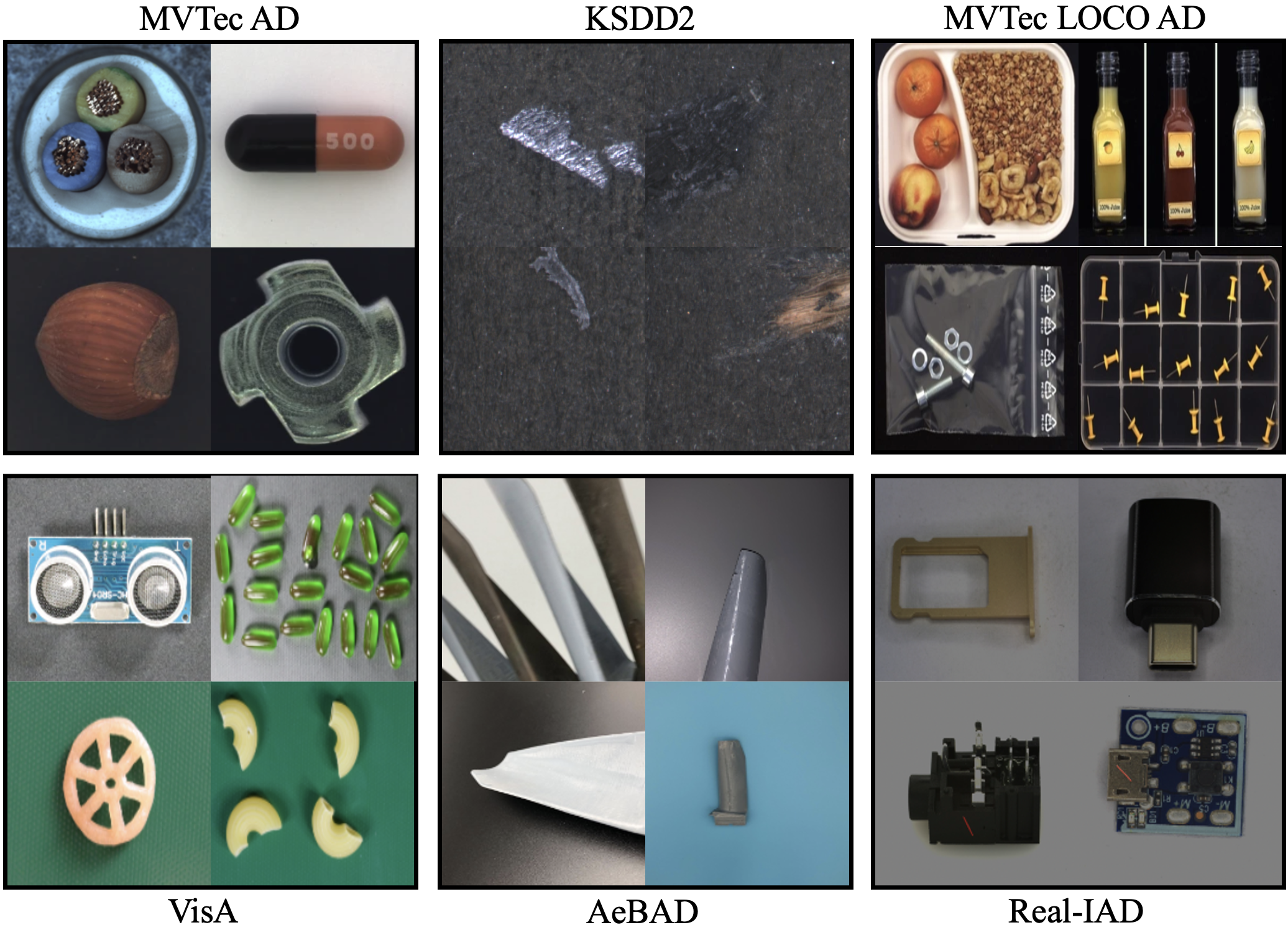}}
\caption{Image dataset examples in defect detection applications. Showcasing a variety of defect types, objects, and class categories.\label{fig11}}
\end{figure}

Numerous datasets have been developed for use in defect detection models in industry. They are used in both the training and testing stages and are crucial to a model’s performance and inspection capabilities. Given that many defect detection applications exist in multiple industries, several datasets can be found. Examples, including prominent ones, are shown in Fig. \ref{fig11}. Additionally, creating reliable datasets is challenging due to the difficulties in collecting, annotating, and classifying large samples of data. Nevertheless, researchers have been able to produce datasets that have gone on to be used in many industrial applications of defect detection. This section presents a summary of the widely utilized datasets in the realm of vision-based IAD. A detailed description of the datasets pertaining to their size, characteristics, and composition is also discussed and collected in Table \ref{table6}.

\begin{table*}[t]
\centering
\caption{Summary and comparison of IAD datasets. Shows a breakdown of the number of images included and their classification.}
\label{table6}
\setlength{\tabcolsep}{3.5pt}
\begin{tabular}{p{40pt}p{50pt}p{30pt}p{350pt}}
\hline
\centering \textbf{Dataset} & \centering \textbf{Non-defective images} & \centering \textbf{Defective images} & \textbf{Description}\\
\hline
\centering MVTec AD (2019) & \centering 3629 & \centering 1725 & USL-based dataset with defect and non-defect high-resolution colored images. Pixel-precise ground truth annotations are provided as well \href{https://www.mvtec.com/company/research/datasets/mvtec-ad}{[link]}.\\
\hline
\centering KSDD (2019)& \centering 347 & \centering 52 & Each of the defect images was annotated with a pixel-wise mask. The images are in grayscale format and have resolutions of 1408×512 pixels \href{https://www.vicos.si/resources/kolektorsdd/}{[link]}.\\
\hline
\centering KSDD2 (2021)& \centering 2979 & \centering 356 & Includes both defective and non-defective images in the training set, with 246 and 2085 images respectively. The testing set has 110 and 894 images respectively \href{https://www.vicos.si/resources/kolektorsdd2/}{[link]}.\\
\hline
\centering MVTec LOCO AD (2022)& \centering 2651 & \centering 993 & Contains 1772 training, 304 validation, and 1568 testing images. The test set contains both non-defective images as well as images with structural and logical anomalies. The number and position of objects in images are fixed to meet logical constraints \href{https://www.mvtec.com/company/research/datasets/mvtec-loco}{[link]}.\\
\hline
\centering VisA \hspace{1cm}(2022)& \centering 9621 & \centering 1200 & Images include location variations, multiple instances, and complex structures. Each class has between 500 and 1000 images and a 4000×6000 high-resolution RGB sensor is used to capture all the images \href{https://github.com/amazon-science/spot-diff}{[link]}.\\
\hline
\centering AeBAD (2023)& \centering 2833 & \centering 2687 & The images are taken in different scales, meaning alignments between images differ. Pixel-wise ground truth annotations are provided for the images and sample-level annotations are provided for the frames in the video dataset \href{https://github.com/zhangzilongc/MMR}{[link]}.\\
\hline
\centering Real-IAD (2024)& \centering 99721 & \centering 51329 & Samples in this dataset include multi-view images that incorporate 5 different angles in the imaging process. A high-resolution camera captures images with a resolution of 3648×5472 \href{https://realiad4ad.github.io/Real-IAD/}{[link]}.\\
\hline
\centering VAD \hspace{1cm}(2024)& \centering 3000 & \centering 2000 & 1000 defective and 2000 non-detective images are included in the training set. In the testing set, 1000 defective and 1000 non-defective images are included. All images have resolutions of 512×512 and are in PNG format \href{https://github.com/abc-125/vad}{[link]}.\\
\hline
\end{tabular}
\end{table*}

The development of datasets can help bridge the gap between research and application. One of the main industrial datasets, MVTec AD \cite{8954181}, was introduced in 2019. It features multiple objects including images of metal nuts, cables, and pills. It is broken into a training and testing bank, with 15 categories and 73 types of anomalies. A total of 1258 defect images are included in the testing bank, as MVTec AD takes an unsupervised approach. Relevant studies applied a student-teacher model for the detection of anomalous features \cite{10484326}. The model was able to achieve an image-level detection AU-ROC of 99.8\% on MVTec AD, indicating its promise for real-world industrial applications. Nonetheless, MVTec AD has limitations in the scarcity of defect images for some categories.

KolektorSDD (KSDD) \cite{Tabernik2019JIM} is a dataset published in 2019 containing images of defective electrical commutators that were annotated by the Kolektor Group. The surfaces of the commutators were captured through eight separate and high-quality images. However, KSDD is limited by its small dataset and focuses on a single defect type, specifically microscopic cracks on a material's surface. KSDD was used in the evaluation of a proposed end-to-end training scheme \cite{9412092}. The results indicated a 100\% detection rate, surpassing previous models. Additionally, they revealed that this dataset has reached saturation.

Given the limitations of KSDD, KolektorSDD2 (KSDD2) \cite{Bozic2021COMIND} was introduced in 2021. It includes several additional defect types, such as surface imperfections and minor spots, which vary in color, size, and shape. Colored images of production items, sized at 230×630 pixels, are captured. Additionally, defects are annotated using pixel-wise segmentation masks. A model implemented on KSDD2 demonstrated the use of a sliding vision transformer to achieve higher accuracy with less annotation labor \cite{li2024}. An AUROC score of 99.1\% was achieved after implementing a semi-supervised approach. However, KSDD2 has not yet been widely adopted due to its limitations in defect sample sizes and single-class.

Similarly, MVTec LOCO AD \cite{locoad} is another unsupervised dataset. It was presented in 2022 and contains 5 categories of objects that include screw bags, pushpins, juice bottles, splicing collectors, and breakfast boxes. The dataset addresses both structural anomalies, such as scratches, and logical anomalies, such as the misplacement of an object. Experiments on a novel part segmentation model, using MVTec LOCO AD, achieved an average AUROC score of 98.1\% in logical anomaly detection \cite{KimPark2024}. However, MVTec LOCO AD has some drawbacks, including its limited object categories and the simplicity of its applications.

VisA \cite{10.1007/978-3-031-20056-4_23} is a popular dataset that is twice as large as MVTec AD. Published in 2022, VisA presents 12 object classes, including PCBs, pill capsules, and macaroni shells, that span 3 domains. Each class has exactly 100 defective images with image and pixel-level annotations being provided. While Visa has a significantly larger dataset, it has fewer classes than MVTec AD, with 4 of its 12 classes being PCBs. The dataset was utilized on a novel defect detection framework with a reconstruction and segmentation sub-network \cite{zhang2023diff}. Results on the 12 classes reflected an average image AUROC score of 98.8\%, outperforming other models on 10 out of the 12 classes.

AeBAD \cite{ZHANG2023103990} is a dataset focused on aero-engine blades. Published in 2023, AeBAD is composed of two datasets: a single-blade dataset and a video inspection dataset of aero blades assembled on a blisk. The training set contains only non-defective samples, whereas the testing set includes four types of blade defects: ablation, groove, breakdown, and fracture. Additionally, a domain shift exists between the testing and training sets. This results in differences in view and illumination of the two sets, allowing for better real-world implementations. The same authors \cite{ZHANG2023103990} tested the AeBAD dataset on a Masked Multi-scale Reconstruction method, which incorporated a feature pyramid network and a hierarchical encoder. AUROC results demonstrated mean scores of 84.7\% and 78.2\% on the image and video datasets respectively, performing better than other state-of-the-art models.

A recent publication has put forward a large industrial anomaly detection dataset to overcome the limitations of currently employed datasets, such as VisA and MVTec. The Real-IAD dataset \cite{wang2024} contains 150 thousand high-resolution images of 30 real-world objects and 8 defect classes. The dataset includes objects such as switches, PCBs, USBs, phone batteries, and plastic nuts. With it being more than ten times larger than other datasets, it aims to set a new point of reference for the IAD field. Experiments demonstrated a performance decrease in image-AUROC from MVTec (97.9\%) to the Real-IAD (85\%) dataset. This suggests that the Real-IAD dataset presents greater challenges for anomaly detection, compared to the current datasets.

The Valeo Anomaly Dataset (VAD) \cite{baitieva2024supervised} was also recently presented. It offers a real-world industrial dataset to fill in the gap of anomaly image scarcity. The images are of a single object, a piezoelectric element, and are collected from a genuine production line. Diverse defects are included, both logical and structural, which showcase the wiring and soundness of the piezoelectric element. Furthermore, images are annotated at the image level, as opposed to the pixel level.

\section{Conclusion and Future Research Directions} 
\label{CONCLUSION AND FUTURE RESEARCH DIRECTIONS}
This survey provides an overview of major concepts in IAD and explores relevant vision-based anomaly detection literature within the context of industrial applications. Emphasis is placed on the level of supervision and prevalent ML and DL models applicable in industrial fields. Aspects of data acquisition, preprocessing, and evaluation are presented and help showcase the general workflow of the IAD process. Relevant industrial datasets are collected and presented as well. The promises of Industry 4.0 and IAD methods are dissected with regards to industrial manufacturing, to point out the benefits and potential challenges this innovation might bring. Lastly, prevalent challenges such as real-time inspection and data imbalance are brought to attention, and avenues for future research directions in the field of vision-based IAD are highlighted. In summary, this review provides researchers with an overview of current IAD motivations, frameworks, and applications, aiming to serve as a reference point for future research in this field. 

Future directions for improving vision-based inspection systems in industry include the continued development of models, datasets, and detection capabilities. One promising research direction is Explainable Artificial Intelligence (XAI). Learning-based models do not explicitly state how they arrive at decisions, making certain outcomes and predictions difficult to understand. The lack of transparency can obscure why some defects are easier to detect than others. XAI aims to enhance the comprehension and interpretability of results from learning algorithms. Explainability seeks to provide researchers with insight into the inner workings of learning models, paving the way for more robust IAD systems. SHapley Additive exPlanations and Local Interpretable Model-Agnostic Explanations are popular techniques designed to improve ML explainability \cite{OLIVEIRA2024110214}. Another promising avenue for vision-based IAD is the use of Large Vision-Language Models (LVLMs). LVLMs are multimodal models, based on Large Language Models (LLMs), that incorporate both images and text into the decision-making process of anomaly detection. They leverage the abstract and rational capabilities of LLMs and expands them into vision-based tasks \cite{10445007}. LVLMs outline some advantages, such as being able to identify defects without the need for threshold settings, in-context learning, and allowing users to ask questions about input images and receive answers \cite{gu2023}. The development and integration of LVLMs will ultimately require loss functions that are able to effectively manage textual data \cite{avideo2024}. Furthermore, as evident by the recent publication of the Real-IAD industrial dataset \cite{wang2024}, the expansion and addition of datasets is a recurring need in this field. As models continue to improve and achieve saturation on available datasets, more complex and challenging datasets are necessary to match the detection requirements of real-world scenarios. Lastly, the continued development of USL, SSL, and transfer learning methods will give more options in the defect detection field by easing the pressure of large, labeled data needs.

\bibliographystyle{IEEEtran}
\bibliography{ref}

\begin{thebibliography}{100}
\providecommand{\url}[1]{#1}
\csname url@samestyle\endcsname
\providecommand{\newblock}{\relax}
\providecommand{\bibinfo}[2]{#2}
\providecommand{\BIBentrySTDinterwordspacing}{\spaceskip=0pt\relax}
\providecommand{\BIBentryALTinterwordstretchfactor}{4}
\providecommand{\BIBentryALTinterwordspacing}{\spaceskip=\fontdimen2\font plus
\BIBentryALTinterwordstretchfactor\fontdimen3\font minus \fontdimen4\font\relax}
\providecommand{\BIBforeignlanguage}[2]{{%
\expandafter\ifx\csname l@#1\endcsname\relax
\typeout{** WARNING: IEEEtran.bst: No hyphenation pattern has been}%
\typeout{** loaded for the language `#1'. Using the pattern for}%
\typeout{** the default language instead.}%
\else
\language=\csname l@#1\endcsname
\fi
#2}}
\providecommand{\BIBdecl}{\relax}
\BIBdecl

\bibitem{articlerev}
P.~Bhatt, R.~Malhan, P.~Rajendran, B.~Shah, S.~Thakar, Y.~J. Yoon, and S.~Gupta, ``Image-based surface defect detection using deep learning: A review,'' \emph{Journal of Computing and Information Science in Engineering}, vol.~21, pp. 1--23, 01 2021.

\bibitem{articlezen}
Z.~Ren, F.~Fang, N.~Yan, and Y.~Wu, ``State of the art in defect detection based on machine vision,'' \emph{International Journal of Precision Engineering and Manufacturing-Green Technology}, vol.~9, p. 661–691, 03 2022.

\bibitem{9900177}
Y.~Tang, K.~Sun, D.~Zhao, Y.~Lu, J.~Jiang, and H.~Chen, ``Industrial defect detection through computer vision: A survey,'' in \emph{2022 7th IEEE International Conference on Data Science in Cyberspace (DSC)}, 2022, pp. 605--610.

\bibitem{LI201866}
\BIBentryALTinterwordspacing
L.~Li, ``China's manufacturing locus in 2025: With a comparison of “made-in-china 2025” and “industry 4.0”,'' \emph{Technological Forecasting and Social Change}, vol. 135, pp. 66--74, 2018. [Online]. Available: \url{https://www.sciencedirect.com/science/article/pii/S0040162517307254}
\BIBentrySTDinterwordspacing

\bibitem{salgues2018society}
B.~Salgues, \emph{Society 5.0: industry of the future, technologies, methods and tools}.\hskip 1em plus 0.5em minus 0.4em\relax John Wiley and Sons, 2018.

\bibitem{QIN20201}
\BIBentryALTinterwordspacing
W.~Qin, S.~Chen, and M.~Peng, ``Recent advances in industrial internet: insights and challenges,'' \emph{Digital Communications and Networks}, vol.~6, no.~1, pp. 1--13, 2020. [Online]. Available: \url{https://www.sciencedirect.com/science/article/pii/S2352864819301166}
\BIBentrySTDinterwordspacing

\bibitem{8843824}
C.~H.~G. Qing, E.~Oktaviani, and S.~Heripracoyo, ``The literature review of manufacturing industry 4.0 using mapping design with gis approach,'' in \emph{2019 International Conference on Information Management and Technology (ICIMTech)}, vol.~1, 2019, pp. 438--443.

\bibitem{ZHENG2024108170}
\BIBentryALTinterwordspacing
H.~Zheng, X.~Chen, H.~Cheng, Y.~Du, and Z.~Jiang, ``Md-yolo: Surface defect detector for industrial complex environments,'' \emph{Optics and Lasers in Engineering}, vol. 178, p. 108170, 2024. [Online]. Available: \url{https://www.sciencedirect.com/science/article/pii/S0143816624001490}
\BIBentrySTDinterwordspacing

\bibitem{doi:10.1080}
\BIBentryALTinterwordspacing
A.~Kusiak, ``Smart manufacturing,'' \emph{International Journal of Production Research}, vol.~56, no. 1-2, pp. 508--517, 2018. [Online]. Available: \url{https://doi.org/10.1080/00207543.2017.1351644}
\BIBentrySTDinterwordspacing

\bibitem{7064655}
J.~Posada, C.~Toro, I.~Barandiaran, D.~Oyarzun, D.~Stricker, R.~de~Amicis, E.~B. Pinto, P.~Eisert, J.~Döllner, and I.~Vallarino, ``Visual computing as a key enabling technology for industrie 4.0 and industrial internet,'' \emph{IEEE Computer Graphics and Applications}, vol.~35, no.~2, pp. 26--40, 2015.

\bibitem{8948233}
Q.~Luo, X.~Fang, L.~Liu, C.~Yang, and Y.~Sun, ``Automated visual defect detection for flat steel surface: A survey,'' \emph{IEEE Transactions on Instrumentation and Measurement}, vol.~69, no.~3, pp. 626--644, 2020.

\bibitem{BERTOLINI2021114820}
\BIBentryALTinterwordspacing
M.~Bertolini, D.~Mezzogori, M.~Neroni, and F.~Zammori, ``Machine learning for industrial applications: A comprehensive literature review,'' \emph{Expert Systems with Applications}, vol. 175, p. 114820, 2021. [Online]. Available: \url{https://www.sciencedirect.com/science/article/pii/S095741742100261X}
\BIBentrySTDinterwordspacing

\bibitem{10113226}
M.~Prunella, R.~M. Scardigno, D.~Buongiorno, A.~Brunetti, N.~Longo, R.~Carli, M.~Dotoli, and V.~Bevilacqua, ``Deep learning for automatic vision-based recognition of industrial surface defects: A survey,'' \emph{IEEE Access}, vol.~11, pp. 43\,370--43\,423, 2023.

\bibitem{GOBERT2018517}
\BIBentryALTinterwordspacing
C.~Gobert, E.~W. Reutzel, J.~Petrich, A.~R. Nassar, and S.~Phoha, ``Application of supervised machine learning for defect detection during metallic powder bed fusion additive manufacturing using high resolution imaging.'' \emph{Additive Manufacturing}, vol.~21, pp. 517--528, 2018. [Online]. Available: \url{https://www.sciencedirect.com/science/article/pii/S2214860417302051}
\BIBentrySTDinterwordspacing

\bibitem{ding2022catching}
C.~Ding, G.~Pang, and C.~Shen, ``Catching both gray and black swans: Open-set supervised anomaly detection,'' in \emph{Proceedings of the IEEE/CVF conference on computer vision and pattern recognition}, 2022, pp. 7388--7398.

\bibitem{artun}
Y.~Guo, S.~Kalinin, H.~Cai, K.~Xiao, S.~Krylyuk, A.~Davydov, Q.~Guo, and A.~Lupini, ``Defect detection in atomic-resolution images via unsupervised learning with translational invariance,'' \emph{npj Computational Materials}, vol.~7, 12 2021.

\bibitem{zabin2023contrastive}
M.~Zabin, A.~N.~B. Kabir, M.~K. Kabir, H.-J. Choi, and J.~Uddin, ``Contrastive self-supervised representation learning framework for metal surface defect detection,'' \emph{Journal of Big Data}, vol.~10, no.~1, p. 145, 2023.

\bibitem{10129970}
Y.~Kim, J.-S. Lee, and J.-H. Lee, ``Automatic defect classification using semi-supervised learning with defect localization,'' \emph{IEEE Transactions on Semiconductor Manufacturing}, vol.~36, no.~3, pp. 476--485, 2023.

\bibitem{shi2024efficient}
C.~Shi, K.~Wang, G.~Zhang, Z.~Li, and C.~Zhu, ``Efficient and accurate semi-supervised semantic segmentation for industrial surface defects,'' \emph{Scientific Reports}, vol.~14, no.~1, p. 21874, 2024.

\bibitem{AHMAD2022181}
\BIBentryALTinterwordspacing
H.~M. Ahmad and A.~Rahimi, ``Deep learning methods for object detection in smart manufacturing: A survey,'' \emph{Journal of Manufacturing Systems}, vol.~64, pp. 181--196, 2022. [Online]. Available: \url{https://www.sciencedirect.com/science/article/pii/S0278612522001066}
\BIBentrySTDinterwordspacing

\bibitem{9849507}
X.~Tao, X.~Gong, X.~Zhang, S.~Yan, and C.~Adak, ``Deep learning for unsupervised anomaly localization in industrial images: A survey,'' \emph{IEEE Transactions on Instrumentation and Measurement}, vol.~71, pp. 1--21, 2022.

\bibitem{YASUDA2022103695}
\BIBentryALTinterwordspacing
Y.~D. Yasuda, F.~A. Cappabianco, L.~E.~G. Martins, and J.~A. Gripp, ``Aircraft visual inspection: A systematic literature review,'' \emph{Computers in Industry}, vol. 141, p. 103695, 2022. [Online]. Available: \url{https://www.sciencedirect.com/science/article/pii/S0166361522000926}
\BIBentrySTDinterwordspacing

\bibitem{10138396}
Y.~Abdulrahman, M.~A.~M. Eltoum, A.~Ayyad, B.~Moyo, and Y.~Zweiri, ``Aero-engine blade defect detection: A systematic review of deep learning models,'' \emph{IEEE Access}, vol.~11, pp. 53\,048--53\,061, 2023.

\bibitem{automanu}
\BIBentryALTinterwordspacing
F.~K. Konstantinidis, N.~Myrillas, K.~A. Tsintotas, S.~G. Mouroutsos, and A.~Gasteratos, ``A technology maturity assessment framework for industry 5.0 machine vision systems based on systematic literature review in automotive manufacturing,'' \emph{International Journal of Production Research}, vol.~0, no.~0, pp. 1--37, 2023. [Online]. Available: \url{https://doi.org/10.1080/00207543.2023.2270588}
\BIBentrySTDinterwordspacing

\bibitem{JHA2023103911}
\BIBentryALTinterwordspacing
S.~B. Jha and R.~F. Babiceanu, ``Deep cnn-based visual defect detection: Survey of current literature,'' \emph{Computers in Industry}, vol. 148, p. 103911, 2023. [Online]. Available: \url{https://www.sciencedirect.com/science/article/pii/S0166361523000611}
\BIBentrySTDinterwordspacing

\bibitem{Liu_2024}
\BIBentryALTinterwordspacing
J.~Liu, G.~Xie, J.~Wang, S.~Li, C.~Wang, F.~Zheng, and Y.~Jin, ``Deep industrial image anomaly detection: A survey,'' \emph{Machine Intelligence Research}, vol.~21, no.~1, p. 104–135, Jan. 2024. [Online]. Available: \url{http://dx.doi.org/10.1007/s11633-023-1459-z}
\BIBentrySTDinterwordspacing

\bibitem{AMERI2024107717}
\BIBentryALTinterwordspacing
R.~Ameri, C.-C. Hsu, and S.~S. Band, ``A systematic review of deep learning approaches for surface defect detection in industrial applications,'' \emph{Engineering Applications of Artificial Intelligence}, vol. 130, p. 107717, 2024. [Online]. Available: \url{https://www.sciencedirect.com/science/article/pii/S0952197623019012}
\BIBentrySTDinterwordspacing

\bibitem{HOJJATI2024106106}
\BIBentryALTinterwordspacing
H.~Hojjati, T.~K.~K. Ho, and N.~Armanfard, ``Self-supervised anomaly detection in computer vision and beyond: A survey and outlook,'' \emph{Neural Networks}, vol. 172, p. 106106, 2024. [Online]. Available: \url{https://www.sciencedirect.com/science/article/pii/S0893608024000200}
\BIBentrySTDinterwordspacing

\bibitem{TRILLES2024101063}
\BIBentryALTinterwordspacing
S.~Trilles, S.~S. Hammad, and D.~Iskandaryan, ``Anomaly detection based on artificial intelligence of things: A systematic literature mapping,'' \emph{Internet of Things}, vol.~25, p. 101063, 2024. [Online]. Available: \url{https://www.sciencedirect.com/science/article/pii/S2542660524000052}
\BIBentrySTDinterwordspacing

\bibitem{electronics13081551}
\BIBentryALTinterwordspacing
E.~Weiss, S.~Caplan, K.~Horn, and M.~Sharabi, ``Real-time defect detection in electronic components during assembly through deep learning,'' \emph{Electronics}, vol.~13, no.~8, 2024. [Online]. Available: \url{https://www.mdpi.com/2079-9292/13/8/1551}
\BIBentrySTDinterwordspacing

\bibitem{10531145}
L.~Yuan, Y.~Chen, H.~Tang, R.~Gao, and W.~Wu, ``A lightweight deep-learning algorithm for welding defect detection in new energy vehicle battery current collectors,'' \emph{IEEE Sensors Journal}, vol.~24, no.~13, pp. 21\,655--21\,668, 2024.

\bibitem{baimachine}
\BIBentryALTinterwordspacing
J.~Bai, D.~Wu, T.~Shelley, P.~Schubel, D.~Twine, J.~Russell, X.~Zeng, and J.~Zhang, ``A comprehensive survey on machine learning driven material defect detection: Challenges, solutions, and future prospects,'' 2024. [Online]. Available: \url{https://arxiv.org/abs/2406.07880}
\BIBentrySTDinterwordspacing

\bibitem{9144}
X.~Dong, C.~Taylor, and T.~Cootes, \emph{\BIBforeignlanguage{English}{Small Defect Detection Using Convolutional Neural Network Features and Random Forests}}.\hskip 1em plus 0.5em minus 0.4em\relax United States: Springer Nature, 2019, vol. 11132, pp. 398--412.

\bibitem{9631303}
S.~Mei, J.~Cheng, X.~He, H.~Hu, and G.~Wen, ``A novel weakly supervised ensemble learning framework for automated pixel-wise industry anomaly detection,'' \emph{IEEE Sensors Journal}, vol.~22, no.~2, pp. 1560--1570, 2022.

\bibitem{li2022bigdatasetgan}
D.~Li, H.~Ling, S.~W. Kim, K.~Kreis, S.~Fidler, and A.~Torralba, ``Bigdatasetgan: Synthesizing imagenet with pixel-wise annotations,'' in \emph{Proceedings of the IEEE/CVF Conference on Computer Vision and Pattern Recognition}, 2022, pp. 21\,330--21\,340.

\bibitem{10500332}
S.~Liu, H.~Ni, C.~Li, Y.~Zou, and Y.~Luo, ``Defectgan: Synthetic data generation for emu defects detection with limited data,'' \emph{IEEE Sensors Journal}, vol.~24, no.~11, pp. 17\,638--17\,652, 2024.

\bibitem{8954051}
Q.~Sun, Y.~Liu, T.-S. Chua, and B.~Schiele, ``Meta-transfer learning for few-shot learning,'' in \emph{2019 IEEE/CVF Conference on Computer Vision and Pattern Recognition (CVPR)}, 2019, pp. 403--412.

\bibitem{PHUYAL2020100023}
\BIBentryALTinterwordspacing
S.~Phuyal, D.~Bista, and R.~Bista, ``Challenges, opportunities and future directions of smart manufacturing: A state of art review,'' \emph{Sustainable Futures}, vol.~2, p. 100023, 2020. [Online]. Available: \url{https://www.sciencedirect.com/science/article/pii/S2666188820300162}
\BIBentrySTDinterwordspacing

\bibitem{10382509}
O.~H. Anidjar, R.~Lang, and M.~Mega, ``Transfer learning methods for fractographic detection of fatigue crack initiation in additive manufacturing,'' \emph{IEEE Access}, vol.~12, pp. 6262--6280, 2024.

\bibitem{YANG2022108338}
\BIBentryALTinterwordspacing
L.~Yang, J.~Fan, B.~Huo, E.~Li, and Y.~Liu, ``A nondestructive automatic defect detection method with pixelwise segmentation,'' \emph{Knowledge-Based Systems}, vol. 242, p. 108338, 2022. [Online]. Available: \url{https://www.sciencedirect.com/science/article/pii/S0950705122001241}
\BIBentrySTDinterwordspacing

\bibitem{9449912}
D.-M. Tsai, S.-K.~S. Fan, and Y.-H. Chou, ``Auto-annotated deep segmentation for surface defect detection,'' \emph{IEEE Transactions on Instrumentation and Measurement}, vol.~70, pp. 1--10, 2021.

\bibitem{ruff2019deep}
L.~Ruff, R.~A. Vandermeulen, N.~G{\"o}rnitz, A.~Binder, E.~M{\"u}ller, K.-R. M{\"u}ller, and M.~Kloft, ``Deep semi-supervised anomaly detection,'' \emph{arXiv preprint arXiv:1906.02694}, 2019.

\bibitem{badmos2020image}
O.~Badmos, A.~Kopp, T.~Bernthaler, and G.~Schneider, ``Image-based defect detection in lithium-ion battery electrode using convolutional neural networks,'' \emph{Journal of Intelligent Manufacturing}, vol.~31, pp. 885--897, 2020.

\bibitem{9537583}
X.~Dong, C.~J. Taylor, and T.~F. Cootes, ``Defect classification and detection using a multitask deep one-class cnn,'' \emph{IEEE Transactions on Automation Science and Engineering}, vol.~19, no.~3, pp. 1719--1730, 2022.

\bibitem{10555503}
L.~Liu, X.~Feng, F.~Li, Q.~Xian, Z.~Chen, and Z.~Jia, ``Surface defect detection of industrial components based on improved yolov5s,'' \emph{IEEE Sensors Journal}, vol.~24, no.~15, pp. 23\,940--23\,950, 2024.

\bibitem{s24010232}
\BIBentryALTinterwordspacing
E.~Cumbajin, N.~Rodrigues, P.~Costa, R.~Miragaia, L.~Frazão, N.~Costa, A.~Fernández-Caballero, J.~Carneiro, L.~H. Buruberri, and A.~Pereira, ``A real-time automated defect detection system for ceramic pieces manufacturing process based on computer vision with deep learning,'' \emph{Sensors}, vol.~24, no.~1, 2024. [Online]. Available: \url{https://www.mdpi.com/1424-8220/24/1/232}
\BIBentrySTDinterwordspacing

\bibitem{9007362}
X.~Fang, W.~Guo, Q.~Li, J.~Zhu, Z.~Chen, J.~Yu, B.~Zhou, and H.~Yang, ``Sewer pipeline fault identification using anomaly detection algorithms on video sequences,'' \emph{IEEE Access}, vol.~8, pp. 39\,574--39\,586, 2020.

\bibitem{9121251}
X.~Zheng, H.~Wang, J.~Chen, Y.~Kong, and S.~Zheng, ``A generic semi-supervised deep learning-based approach for automated surface inspection,'' \emph{IEEE Access}, vol.~8, pp. 114\,088--114\,099, 2020.

\bibitem{LI2023102470}
\BIBentryALTinterwordspacing
W.~Li, H.~Zhang, G.~Wang, G.~Xiong, M.~Zhao, G.~Li, and R.~Li, ``Deep learning based online metallic surface defect detection method for wire and arc additive manufacturing,'' \emph{Robotics and Computer-Integrated Manufacturing}, vol.~80, p. 102470, 2023. [Online]. Available: \url{https://www.sciencedirect.com/science/article/pii/S0736584522001521}
\BIBentrySTDinterwordspacing

\bibitem{BABIC2021262}
\BIBentryALTinterwordspacing
M.~Babic, M.~A. Farahani, and T.~Wuest, ``Image based quality inspection in smart manufacturing systems: A literature review,'' \emph{Procedia CIRP}, vol. 103, pp. 262--267, 2021, 9th CIRP Global Web Conference – Sustainable, resilient, and agile manufacturing and service operations : Lessons from COVID-19. [Online]. Available: \url{https://www.sciencedirect.com/science/article/pii/S2212827121008830}
\BIBentrySTDinterwordspacing

\bibitem{arviso}
D.~Prasad, M.~Sreekumar, and P.~Karumbu, ``Identification and classification of materials using machine vision and machine learning in the context of industry 4.0,'' \emph{Journal of Intelligent Manufacturing}, vol.~31, 06 2020.

\bibitem{ainspec}
M.~Ali and A.~Lun, ``A cascading fuzzy logic with image processing algorithm–based defect detection for automatic visual inspection of industrial cylindrical object’s surface,'' \emph{The International Journal of Advanced Manufacturing Technology}, vol. 102, 05 2019.

\bibitem{abdulrahman2022ai}
Y.~Abdulrahman, V.~Parezanovic, and D.~Svetinovic, ``Ai-blockchain systems in aerospace engineering and management: Review and challenges,'' in \emph{2022 30th Telecommunications Forum (TELFOR)}.\hskip 1em plus 0.5em minus 0.4em\relax IEEE, 2022, pp. 1--4.

\bibitem{electronics11152383}
\BIBentryALTinterwordspacing
Z.~Chen, J.~Deng, Q.~Zhu, H.~Wang, and Y.~Chen, ``A systematic review of machine-vision-based leather surface defect inspection,'' \emph{Electronics}, vol.~11, no.~15, 2022. [Online]. Available: \url{https://www.mdpi.com/2079-9292/11/15/2383}
\BIBentrySTDinterwordspacing

\bibitem{mulview}
T.~Kaichi, S.~Mori, H.~Saito, J.~Sugano, and H.~Adachi, ``Multi-view surface inspection using a rotating table,'' \emph{Electronic Imaging}, vol. 2018, pp. 1--5, 01 2018.

\bibitem{carview}
R.~Ruitenbeek and S.~Bhulai, ``Multi-view damage inspection using single-view damage projection,'' \emph{Machine Vision and Applications}, vol.~33, 05 2022.

\bibitem{10.1007/978-1-4471-1580-9_18}
S.~K. Nayar, ``Omnidirectional vision,'' in \emph{Robotics Research}, Y.~Shirai and S.~Hirose, Eds.\hskip 1em plus 0.5em minus 0.4em\relax London: Springer London, 1998, pp. 195--202.

\bibitem{ZHANG2024108003}
\BIBentryALTinterwordspacing
P.~Zhang, F.~Zhou, X.~Wang, S.~Wang, and Z.~Song, ``Omnidirectional imaging sensor based on conical mirror for pipelines,'' \emph{Optics and Lasers in Engineering}, vol. 175, p. 108003, 2024. [Online]. Available: \url{https://www.sciencedirect.com/science/article/pii/S0143816623005328}
\BIBentrySTDinterwordspacing

\bibitem{s23094444}
\BIBentryALTinterwordspacing
Q.~Fang, C.~Ibarra-Castanedo, I.~Garrido, Y.~Duan, and X.~Maldague, ``Automatic detection and identification of defects by deep learning algorithms from pulsed thermography data,'' \emph{Sensors}, vol.~23, no.~9, 2023. [Online]. Available: \url{https://www.mdpi.com/1424-8220/23/9/4444}
\BIBentrySTDinterwordspacing

\bibitem{s23218780}
\BIBentryALTinterwordspacing
X.~Zhao, Y.~Zhao, S.~Hu, H.~Wang, Y.~Zhang, and W.~Ming, ``Progress in active infrared imaging for defect detection in the renewable and electronic industries,'' \emph{Sensors}, vol.~23, no.~21, 2023. [Online]. Available: \url{https://www.mdpi.com/1424-8220/23/21/8780}
\BIBentrySTDinterwordspacing

\bibitem{pvarticle}
C.~Buerhop, L.~Bommes, J.~Schlipf, T.~Pickel, A.~Fladung, and I.~M. Peters, ``Infrared imaging of photovoltaic modules a review of the state of the art and future challenges facing gigawatt photovoltaic power stations,'' \emph{Progress in Energy}, vol.~4, 08 2022.

\bibitem{WANG2023112307}
\BIBentryALTinterwordspacing
F.~Wang, Y.~Zhou, X.~Zhang, Z.~Li, J.~Weng, G.~Qiang, M.~Chen, Y.~Wang, H.~Yue, and J.~Liu, ``Laser-induced thermography: An effective detection approach for multiple-type defects of printed circuit boards (pcbs) multilayer complex structure,'' \emph{Measurement}, vol. 206, p. 112307, 2023. [Online]. Available: \url{https://www.sciencedirect.com/science/article/pii/S0263224122015032}
\BIBentrySTDinterwordspacing

\bibitem{noauthor_need_2018}
\BIBentryALTinterwordspacing
J.~Hudson, ``The need for precision lighting control in machine vision,'' 05 2018. [Online]. Available: \url{https://www.novuslight.com/the-need-for-precision-lighting-control-in-machine-vision_N8017.html}
\BIBentrySTDinterwordspacing

\bibitem{noauthor_3d}
\BIBentryALTinterwordspacing
``3d vision: Mvtec software.'' [Online]. Available: \url{https://www.mvtec.com/technologies/3d-vision}
\BIBentrySTDinterwordspacing

\bibitem{s18020408}
\BIBentryALTinterwordspacing
S.~Barone, M.~Carulli, P.~Neri, A.~Paoli, and A.~V. Razionale, ``An omnidirectional vision sensor based on a spherical mirror catadioptric system,'' \emph{Sensors}, vol.~18, no.~2, 2018. [Online]. Available: \url{https://www.mdpi.com/1424-8220/18/2/408}
\BIBentrySTDinterwordspacing

\bibitem{BAGAVATHIAPPAN201335}
\BIBentryALTinterwordspacing
S.~Bagavathiappan, B.~Lahiri, T.~Saravanan, J.~Philip, and T.~Jayakumar, ``Infrared thermography for condition monitoring – a review,'' \emph{Infrared Physics and Technology}, vol.~60, pp. 35--55, 2013. [Online]. Available: \url{https://www.sciencedirect.com/science/article/pii/S1350449513000327}
\BIBentrySTDinterwordspacing

\bibitem{5445596}
C.~Saravanan, ``Color image to grayscale image conversion,'' in \emph{2010 Second International Conference on Computer Engineering and Applications}, vol.~2, 2010, pp. 196--199.

\bibitem{10311554}
Y.~Zheng, S.~Li, Y.~Xiang, and Z.~Zhu, ``Crack defect detection processing algorithm and method of mems devices based on image processing technology,'' \emph{IEEE Access}, vol.~11, pp. 126\,323--126\,334, 2023.

\bibitem{10466741}
S.~Ahmed, J.~F. Esha, M.~S. Rahman, M.~S. Kaiser, A.~S. M.~S. Hosen, D.~Ghimire, and M.~J. Park, ``Exploring deep learning and machine learning approaches for brain hemorrhage detection,'' \emph{IEEE Access}, vol.~12, pp. 45\,060--45\,093, 2024.

\bibitem{NG20061644}
\BIBentryALTinterwordspacing
H.-F. Ng, ``Automatic thresholding for defect detection,'' \emph{Pattern Recognition Letters}, vol.~27, no.~14, pp. 1644--1649, 2006. [Online]. Available: \url{https://www.sciencedirect.com/science/article/pii/S016786550600119X}
\BIBentrySTDinterwordspacing

\bibitem{UESUGI2023103442}
\BIBentryALTinterwordspacing
F.~Uesugi, ``Novel image processing method inspired by wavelet transform,'' \emph{Micron}, vol. 168, p. 103442, 2023. [Online]. Available: \url{https://www.sciencedirect.com/science/article/pii/S0968432823000409}
\BIBentrySTDinterwordspacing

\bibitem{CERVANTES2020189}
\BIBentryALTinterwordspacing
J.~Cervantes, F.~Garcia-Lamont, L.~Rodríguez-Mazahua, and A.~Lopez, ``A comprehensive survey on support vector machine classification: Applications, challenges and trends,'' \emph{Neurocomputing}, vol. 408, pp. 189--215, 2020. [Online]. Available: \url{https://www.sciencedirect.com/science/article/pii/S0925231220307153}
\BIBentrySTDinterwordspacing

\bibitem{9065747}
K.~Taunk, S.~De, S.~Verma, and A.~Swetapadma, ``A brief review of nearest neighbor algorithm for learning and classification,'' in \emph{2019 International Conference on Intelligent Computing and Control Systems (ICCS)}, 2019, pp. 1255--1260.

\bibitem{jourNB}
I.~Wickramasinghe and H.~Kalutarage, ``Naive bayes: applications, variations and vulnerabilities: a review of literature with code snippets for implementation indika,'' \emph{Soft Computing}, vol.~24, 02 2021.

\bibitem{9918857}
Y.~Lu, T.~Ye, and J.~Zheng, ``Decision tree algorithm in machine learning,'' in \emph{2022 IEEE International Conference on Advances in Electrical Engineering and Computer Applications (AEECA)}, 2022, pp. 1014--1017.

\bibitem{breiman_random_2001}
\BIBentryALTinterwordspacing
L.~Breiman, ``Random {Forests},'' \emph{Machine Learning}, vol.~45, no.~1, pp. 5--32, Oct. 2001. [Online]. Available: \url{https://doi.org/10.1023/A:1010933404324}
\BIBentrySTDinterwordspacing

\bibitem{defCNN}
R.~Yamashita, M.~Nishio, R.~Do, and K.~Togashi, ``Convolutional neural networks: an overview and application in radiology,'' \emph{Insights into Imaging}, vol.~9, 06 2018.

\bibitem{articlegancite}
X.~He, Z.~Chang, L.~Zhang, H.~Xu, H.~Chen, and Z.~Luo, ``A survey of defect detection applications based on generative adversarial networks,'' \emph{IEEE Access}, vol.~10, pp. 113\,493--113\,512, 2022.

\bibitem{7780460}
J.~Redmon, S.~Divvala, R.~Girshick, and A.~Farhadi, ``You only look once: Unified, real-time object detection,'' in \emph{2016 IEEE Conference on Computer Vision and Pattern Recognition (CVPR)}, 2016, pp. 779--788.

\bibitem{JIANG20221066}
\BIBentryALTinterwordspacing
P.~Jiang, D.~Ergu, F.~Liu, Y.~Cai, and B.~Ma, ``A review of yolo algorithm developments,'' \emph{Procedia Computer Science}, vol. 199, pp. 1066--1073, 2022, the 8th International Conference on Information Technology and Quantitative Management (ITQM 2020 and 2021): Developing Global Digital Economy after COVID-19. [Online]. Available: \url{https://www.sciencedirect.com/science/article/pii/S1877050922001363}
\BIBentrySTDinterwordspacing

\bibitem{10539065}
W.~Yaodong, Y.~Hang, G.~Baoqing, S.~Hongmei, and Y.~Zujun, ``Research on real-time detection system of rail surface defects based on deep learning,'' \emph{IEEE Sensors Journal}, vol.~24, no.~13, pp. 21\,157--21\,167, 2024.

\bibitem{9875035}
S.~Tian, P.~Huang, H.~Ma, J.~Wang, X.~Zhou, S.~Zhang, J.~Zhou, R.~Huang, and Y.~Li, ``Casdd: Automatic surface defect detection using a complementary adversarial network,'' \emph{IEEE Sensors Journal}, vol.~22, no.~20, pp. 19\,583--19\,595, 2022.

\bibitem{7780459}
K.~He, X.~Zhang, S.~Ren, and J.~Sun, ``Deep residual learning for image recognition,'' in \emph{2016 IEEE Conference on Computer Vision and Pattern Recognition (CVPR)}, 2016, pp. 770--778.

\bibitem{Parnami2}
\BIBentryALTinterwordspacing
A.~Parnami and M.~Lee, ``Learning from few examples: A summary of approaches to few-shot learning,'' \emph{ArXiv}, vol. abs/2203.04291, 2022. [Online]. Available: \url{https://api.semanticscholar.org/CorpusID:247318847}
\BIBentrySTDinterwordspacing

\bibitem{9623623}
H.~Wang, Z.~Li, and H.~Wang, ``Few-shot steel surface defect detection,'' \emph{IEEE Transactions on Instrumentation and Measurement}, vol.~71, pp. 1--12, 2022.

\bibitem{10204096}
J.~Jeong, Y.~Zou, T.~Kim, D.~Zhang, A.~Ravichandran, and O.~Dabeer, ``Winclip: Zero-/few-shot anomaly classification and segmentation,'' in \emph{2023 IEEE/CVF Conference on Computer Vision and Pattern Recognition (CVPR)}, 2023, pp. 19\,606--19\,616.

\bibitem{radford}
\BIBentryALTinterwordspacing
A.~Radford, J.~W. Kim, C.~Hallacy, A.~Ramesh, G.~Goh, S.~Agarwal, G.~Sastry, A.~Askell, P.~Mishkin, J.~Clark, G.~Krueger, and I.~Sutskever, ``Learning transferable visual models from natural language supervision,'' 2021. [Online]. Available: \url{https://arxiv.org/abs/2103.00020}
\BIBentrySTDinterwordspacing

\bibitem{9879738}
K.~Roth, L.~Pemula, J.~Zepeda, B.~Schölkopf, T.~Brox, and P.~Gehler, ``Towards total recall in industrial anomaly detection,'' in \emph{2022 IEEE/CVF Conference on Computer Vision and Pattern Recognition (CVPR)}, 2022, pp. 14\,298--14\,308.

\bibitem{Hinton2015DistillingTK}
\BIBentryALTinterwordspacing
G.~E. Hinton, O.~Vinyals, and J.~Dean, ``Distilling the knowledge in a neural network,'' \emph{ArXiv}, vol. abs/1503.02531, 2015. [Online]. Available: \url{https://api.semanticscholar.org/CorpusID:7200347}
\BIBentrySTDinterwordspacing

\bibitem{hu2023teacher}
\BIBentryALTinterwordspacing
C.~Hu, X.~Li, D.~Liu, H.~Wu, X.~Chen, J.~Wang, and X.~Liu, ``Teacher-student architecture for knowledge distillation: A survey,'' 2023. [Online]. Available: \url{https://arxiv.org/abs/2308.04268}
\BIBentrySTDinterwordspacing

\bibitem{9157778}
P.~Bergmann, M.~Fauser, D.~Sattlegger, and C.~Steger, ``Uninformed students: Student-teacher anomaly detection with discriminative latent embeddings,'' in \emph{2020 IEEE/CVF Conference on Computer Vision and Pattern Recognition (CVPR)}, 2020, pp. 4182--4191.

\bibitem{jiang2022softpatch}
X.~Jiang, J.~Liu, J.~Wang, Q.~Nie, K.~Wu, Y.~Liu, C.~Wang, and F.~Zheng, ``Softpatch: Unsupervised anomaly detection with noisy data,'' \emph{Advances in Neural Information Processing Systems}, vol.~35, pp. 15\,433--15\,445, 2022.

\bibitem{9306873}
Y.-T. Li, P.~Kuo, and J.-I. Guo, ``Automatic industry pcb board dip process defect detection system based on deep ensemble self-adaption method,'' \emph{IEEE Transactions on Components, Packaging and Manufacturing Technology}, vol.~11, no.~2, pp. 312--323, 2021.

\bibitem{10472782}
Y.~Chu, Y.~He, X.~Xiong, Y.~Lou, C.~Yu, and L.~Duan, ``Augmented hybrid learning for visual defect inspection in real-world hydrogen storage manufacturing scenarios,'' \emph{IEEE Transactions on Industrial Informatics}, vol.~20, no.~6, pp. 8477--8487, 2024.

\bibitem{9530369}
D.~Cannizzaro, A.~G. Varrella, S.~Paradiso, R.~Sampieri, Y.~Chen, A.~Macii, E.~Patti, and S.~D. Cataldo, ``In-situ defect detection of metal additive manufacturing: An integrated framework,'' \emph{IEEE Transactions on Emerging Topics in Computing}, vol.~10, no.~1, pp. 74--86, 2022.

\bibitem{Tabernik2019JIM}
D.~Tabernik, S.~{\v{S}}ela, J.~Skvar{\v{c}}, and D.~Sko{\v{c}}aj, ``Segmentation-based deep-learning approach for surface-defect detection,'' \emph{Journal of Intelligent Manufacturing}, 05 2019.

\bibitem{bugatti2022towards}
M.~Bugatti and B.~M. Colosimo, ``Towards real-time in-situ monitoring of hot-spot defects in l-pbf: A new classification-based method for fast video-imaging data analysis,'' \emph{Journal of Intelligent Manufacturing}, vol.~33, no.~1, pp. 293--309, 2022.

\bibitem{MUJEEB2019100933}
\BIBentryALTinterwordspacing
A.~Mujeeb, W.~Dai, M.~Erdt, and A.~Sourin, ``One class based feature learning approach for defect detection using deep autoencoders,'' \emph{Advanced Engineering Informatics}, vol.~42, p. 100933, 2019. [Online]. Available: \url{https://www.sciencedirect.com/science/article/pii/S1474034619301259}
\BIBentrySTDinterwordspacing

\bibitem{XIA2020845}
\BIBentryALTinterwordspacing
C.~Xia, Z.~Pan, Z.~Fei, S.~Zhang, and H.~Li, ``Vision based defects detection for keyhole tig welding using deep learning with visual explanation,'' \emph{Journal of Manufacturing Processes}, vol.~56, pp. 845--855, 2020. [Online]. Available: \url{https://www.sciencedirect.com/science/article/pii/S1526612520303480}
\BIBentrySTDinterwordspacing

\bibitem{davtalab2022automated}
O.~Davtalab, A.~Kazemian, X.~Yuan, and B.~Khoshnevis, ``Automated inspection in robotic additive manufacturing using deep learning for layer deformation detection,'' \emph{Journal of Intelligent Manufacturing}, vol.~33, no.~3, pp. 771--784, 2022.

\bibitem{zhu2020vision}
J.~Zhu, C.~Zhang, H.~Qi, and Z.~Lu, ``Vision-based defects detection for bridges using transfer learning and convolutional neural networks,'' \emph{Structure and Infrastructure Engineering}, vol.~16, no.~7, pp. 1037--1049, 2020.

\bibitem{MENDOZABERNAL2024123210}
\BIBentryALTinterwordspacing
J.~Mendoza-Bernal, A.~González-Vidal, and A.~F. Skarmeta, ``A convolutional neural network approach for image-based anomaly detection in smart agriculture,'' \emph{Expert Systems with Applications}, vol. 247, p. 123210, 2024. [Online]. Available: \url{https://www.sciencedirect.com/science/article/pii/S0957417424000757}
\BIBentrySTDinterwordspacing

\bibitem{amieghemen2023deep}
G.~E. Amieghemen and M.~M. Sherif, ``Deep convolutional neural network ensemble for pavement crack detection using high elevation uav images,'' \emph{Structure and Infrastructure Engineering}, pp. 1--16, 2023.

\bibitem{huang2022deep}
H.~Huang, S.~Zhao, D.~Zhang, and J.~Chen, ``Deep learning-based instance segmentation of cracks from shield tunnel lining images,'' \emph{Structure and Infrastructure Engineering}, vol.~18, no.~2, pp. 183--196, 2022.

\bibitem{s21154968}
\BIBentryALTinterwordspacing
J.~Kim, J.~Ko, H.~Choi, and H.~Kim, ``Printed circuit board defect detection using deep learning via a skip-connected convolutional autoencoder,'' \emph{Sensors}, vol.~21, no.~15, 2021. [Online]. Available: \url{https://www.mdpi.com/1424-8220/21/15/4968}
\BIBentrySTDinterwordspacing

\bibitem{chen2020robust}
H.~Chen, Q.~Hu, B.~Zhai, H.~Chen, and K.~Liu, ``A robust weakly supervised learning of deep conv-nets for surface defect inspection,'' \emph{Neural Computing and Applications}, vol.~32, pp. 11\,229--11\,244, 2020.

\bibitem{iqbal2024multi}
E.~Iqbal, S.~U. Khan, S.~Javed, B.~Moyo, Y.~Zweiri, and Y.~Abdulrahman, ``Multi-scale feature reconstruction network for industrial anomaly detection,'' \emph{Knowledge-Based Systems}, vol. 305, p. 112650, 2024.

\bibitem{9736943}
J.~Tao, Y.~Zhu, F.~Jiang, H.~Liu, and H.~Liu, ``Rolling surface defect inspection for drum-shaped rollers based on deep learning,'' \emph{IEEE Sensors Journal}, vol.~22, no.~9, pp. 8693--8700, 2022.

\bibitem{zhang2023automated}
J.~Zhang, S.~Qian, and C.~Tan, ``Automated bridge crack detection method based on lightweight vision models,'' \emph{Complex \& Intelligent Systems}, vol.~9, pp. 1--14, 09 2022.

\bibitem{10416966}
X.~Li, H.~Yan, K.~Cui, Z.~Li, R.~Liu, G.~Lu, K.~C. Hsieh, X.~Liu, and C.~Hon, ``A novel hybrid yolo approach for precise paper defect detection with a dual-layer template and an attention mechanism,'' \emph{IEEE Sensors Journal}, vol.~24, no.~7, pp. 11\,651--11\,669, 2024.

\bibitem{liong2020leather}
S.-T. Liong, D.~Zheng, Y.-C. Huang, and Y.~S. Gan, ``Leather defect classification and segmentation using deep learning architecture,'' \emph{International Journal of Computer Integrated Manufacturing}, vol.~33, no. 10-11, pp. 1105--1117, 2020.

\bibitem{XU2024124244}
\BIBentryALTinterwordspacing
G.~Xu, M.~Ren, and G.~Li, ``Efficient online surface defect detection using multiple instance learning,'' \emph{Expert Systems with Applications}, vol. 252, p. 124244, 2024. [Online]. Available: \url{https://www.sciencedirect.com/science/article/pii/S0957417424011102}
\BIBentrySTDinterwordspacing

\bibitem{8954181}
P.~Bergmann, M.~Fauser, D.~Sattlegger, and C.~Steger, ``Mvtec ad — a comprehensive real-world dataset for unsupervised anomaly detection,'' in \emph{2019 IEEE/CVF Conference on Computer Vision and Pattern Recognition (CVPR)}, 2019, pp. 9584--9592.

\bibitem{10484326}
K.~Batzner, L.~Heckler, and R.~König, ``Efficientad: Accurate visual anomaly detection at millisecond-level latencies,'' in \emph{2024 IEEE/CVF Winter Conference on Applications of Computer Vision (WACV)}, 2024, pp. 127--137.

\bibitem{9412092}
J.~Božič, D.~Tabernik, and D.~Skočaj, ``End-to-end training of a two-stage neural network for defect detection,'' in \emph{2020 25th International Conference on Pattern Recognition (ICPR)}, 2021, pp. 5619--5626.

\bibitem{Bozic2021COMIND}
J.~Bo{\v{z}}i{\v{c}}, D.~Tabernik, and D.~Sko{\v{c}}aj, ``{Mixed supervision for surface-defect detection: from weakly to fully supervised learning},'' \emph{Computers in Industry}, 2021.

\bibitem{li2024}
\BIBentryALTinterwordspacing
H.~Li, J.~Wu, H.~Chen, M.~Wang, and C.~Shen, ``Efficient anomaly detection with budget annotation using semi-supervised residual transformer,'' 2024. [Online]. Available: \url{https://arxiv.org/abs/2306.03492}
\BIBentrySTDinterwordspacing

\bibitem{locoad}
P.~Bergmann, K.~Batzner, M.~Fauser, D.~Sattlegger, and C.~Steger, ``Beyond dents and scratches: Logical constraints in unsupervised anomaly detection and localization,'' \emph{International Journal of Computer Vision}, vol. 130, 04 2022.

\bibitem{KimPark2024}
\BIBentryALTinterwordspacing
S.~Kim, S.~An, P.~Chikontwe, M.~Kang, E.~Adeli, K.~M. Pohl, and S.~H. Park, ``Few shot part segmentation reveals compositional logic for industrial anomaly detection,'' 2024. [Online]. Available: \url{https://arxiv.org/abs/2312.13783}
\BIBentrySTDinterwordspacing

\bibitem{10.1007/978-3-031-20056-4_23}
Y.~Zou, J.~Jeong, L.~Pemula, D.~Zhang, and O.~Dabeer, ``Spot-the-difference self-supervised pre-training for anomaly detection and segmentation,'' in \emph{Computer Vision -- ECCV 2022}, S.~Avidan, G.~Brostow, M.~Ciss{\'e}, G.~M. Farinella, and T.~Hassner, Eds.\hskip 1em plus 0.5em minus 0.4em\relax Cham: Springer Nature Switzerland, 2022, pp. 392--408.

\bibitem{zhang2023diff}
\BIBentryALTinterwordspacing
H.~Zhang, Z.~Wang, Z.~Wu, and Y.-G. Jiang, ``Diffusionad: Norm-guided one-step denoising diffusion for anomaly detection,'' 2023. [Online]. Available: \url{https://arxiv.org/abs/2303.08730}
\BIBentrySTDinterwordspacing

\bibitem{ZHANG2023103990}
\BIBentryALTinterwordspacing
Z.~Zhang, Z.~Zhao, X.~Zhang, C.~Sun, and X.~Chen, ``Industrial anomaly detection with domain shift: A real-world dataset and masked multi-scale reconstruction,'' \emph{Computers in Industry}, vol. 151, p. 103990, 2023. [Online]. Available: \url{https://www.sciencedirect.com/science/article/pii/S0166361523001409}
\BIBentrySTDinterwordspacing

\bibitem{wang2024}
\BIBentryALTinterwordspacing
C.~Wang, W.~Zhu, B.-B. Gao, Z.~Gan, J.~Zhang, Z.~Gu, S.~Qian, M.~Chen, and L.~Ma, ``Real-iad: A real-world multi-view dataset for benchmarking versatile industrial anomaly detection,'' 2024. [Online]. Available: \url{https://arxiv.org/abs/2403.12580}
\BIBentrySTDinterwordspacing

\bibitem{baitieva2024supervised}
A.~Baitieva, D.~Hurych, V.~Besnier, and O.~Bernard, ``Supervised anomaly detection for complex industrial images,'' 2024.

\bibitem{OLIVEIRA2024110214}
\BIBentryALTinterwordspacing
R.~M.~A. Oliveira, Ângelo Márcio Oliveira~Sant’Anna, and P.~H.~F. {da Silva}, ``Explainable machine learning models for defects detection in industrial processes,'' \emph{Computers and Industrial Engineering}, vol. 192, p. 110214, 2024. [Online]. Available: \url{https://www.sciencedirect.com/science/article/pii/S0360835224003358}
\BIBentrySTDinterwordspacing

\bibitem{10445007}
J.~Zhang, J.~Huang, S.~Jin, and S.~Lu, ``Vision-language models for vision tasks: A survey,'' \emph{IEEE Transactions on Pattern Analysis and Machine Intelligence}, vol.~46, no.~8, pp. 5625--5644, 2024.

\bibitem{gu2023}
\BIBentryALTinterwordspacing
Z.~Gu, B.~Zhu, G.~Zhu, Y.~Chen, M.~Tang, and J.~Wang, ``Anomalygpt: Detecting industrial anomalies using large vision-language models,'' 2023. [Online]. Available: \url{https://arxiv.org/abs/2308.15366}
\BIBentrySTDinterwordspacing

\bibitem{avideo2024}
\BIBentryALTinterwordspacing
M.~Abdalla, S.~Javed, M.~A. Radi, A.~Ulhaq, and N.~Werghi, ``Video anomaly detection in 10 years: A survey and outlook,'' 2024. [Online]. Available: \url{https://arxiv.org/abs/2405.19387}
\BIBentrySTDinterwordspacing

\end{thebibliography}

\begin{IEEEbiography}[{\includegraphics[width=1in,height=1.25in,clip,keepaspectratio]{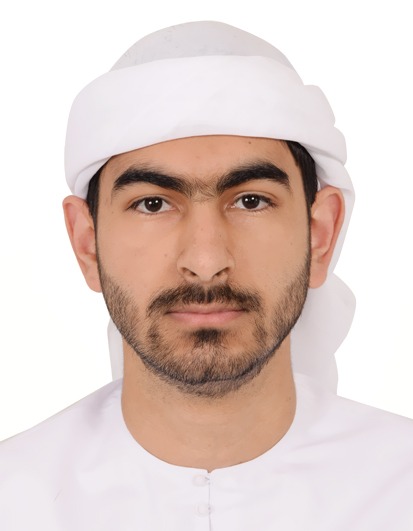}}]{Abdelrahman Alzarooni} received the B.S. degree in aerospace engineering from the Pennsylvania State University, State College, PA, USA, in 2023. He is currently pursuing the Ph.D. degree with Khalifa University of Science and Technology, Abu Dhabi, UAE.

His research interests include automated defect detection, aero-engine blade inspection, and non-destructive testing techniques.
\end{IEEEbiography}

\begin{IEEEbiography}[{\includegraphics[width=1in,height=1.25in,clip,keepaspectratio]{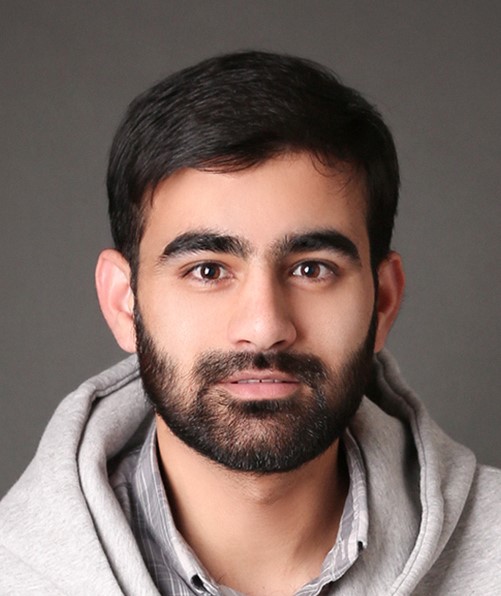}}]{Ehtesham Iqbal} received his B.S. degree in electrical (computer) engineering from the COMSATS University, Islamabad, Pakistan, in 2017, and the M.S. degree in computer science and engineering from Chung-Ang University, Seoul, South Korea, in 2021.

He worked as a Research Assistant at the Visual Image Media Laboratory in Seoul, South Korea. Later, he worked as a deep learning engineer at AiV Research Group South Korea. He is a research associate at the Advanced Research and Innovation Center (ARIC), Khalifa University of Science and Technology, Abu Dhabi, UAE. His current research interests include industrial anomaly detection, few-shot Learning, and Optical Character Recognition. 
\end{IEEEbiography}

\begin{IEEEbiography}[{\includegraphics[width=1in,height=1.25in,clip,keepaspectratio]{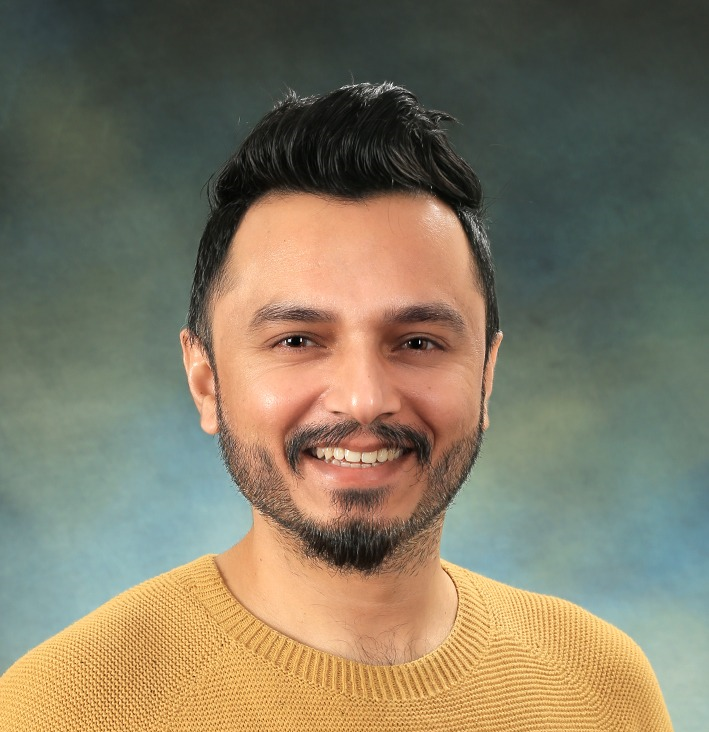}}]{Samee U. Khan} (Member, IEEE) received the Ph.D. degree in software convergence from Sejong University, Seoul, South Korea, in 2023. He is currently a Postdoctoral Fellow Advanced Research and Innovation Center (ARIC), Khalifa University of Science and Technology, Abu Dhabi, UAE. His main research interests include multimedia data analysis and multimodal, including person re-identification, industrial anomaly, human activity recognition, video summarization, and time series data analysis for power generation and consumption prediction/forecasting. He has published several articles in peer-reviewed journals in reputed venues, including IEEE TRANSACTIONS ON INDUSTRIAL INFORMATICS, IEEE. INTERNET OF THINGS JOURNAL, IEEE MULTIMEDIA, IEEE JOURNAL OF SELECTED. TOPICS IN SIGNAL PROCESSING, IEEE Access, Energy and Buildings, Alexandria Engineering Journal, Renewable and Sustainable Energy Reviews, International Journal of Intelligent Systems, and Multimedia Tools and Applications (Springer). He provides professional review services in various reputed journals, including IEEE INTERNET OF THINGS JOURNAL, IEEE OPEN JOURNAL OF ENGINEERING IN MEDICINE AND BIOLOGY, IEEE TRANSACTIONS ON GREEN COMMUNICATIONS AND NETWORKING, IEEE TRANSACTIONS ON HUMAN-MACHINE SYSTEMS, and ASCJ.

\end{IEEEbiography}

\begin{IEEEbiography}[{\includegraphics[width=1in,height=1.25in,clip,keepaspectratio]{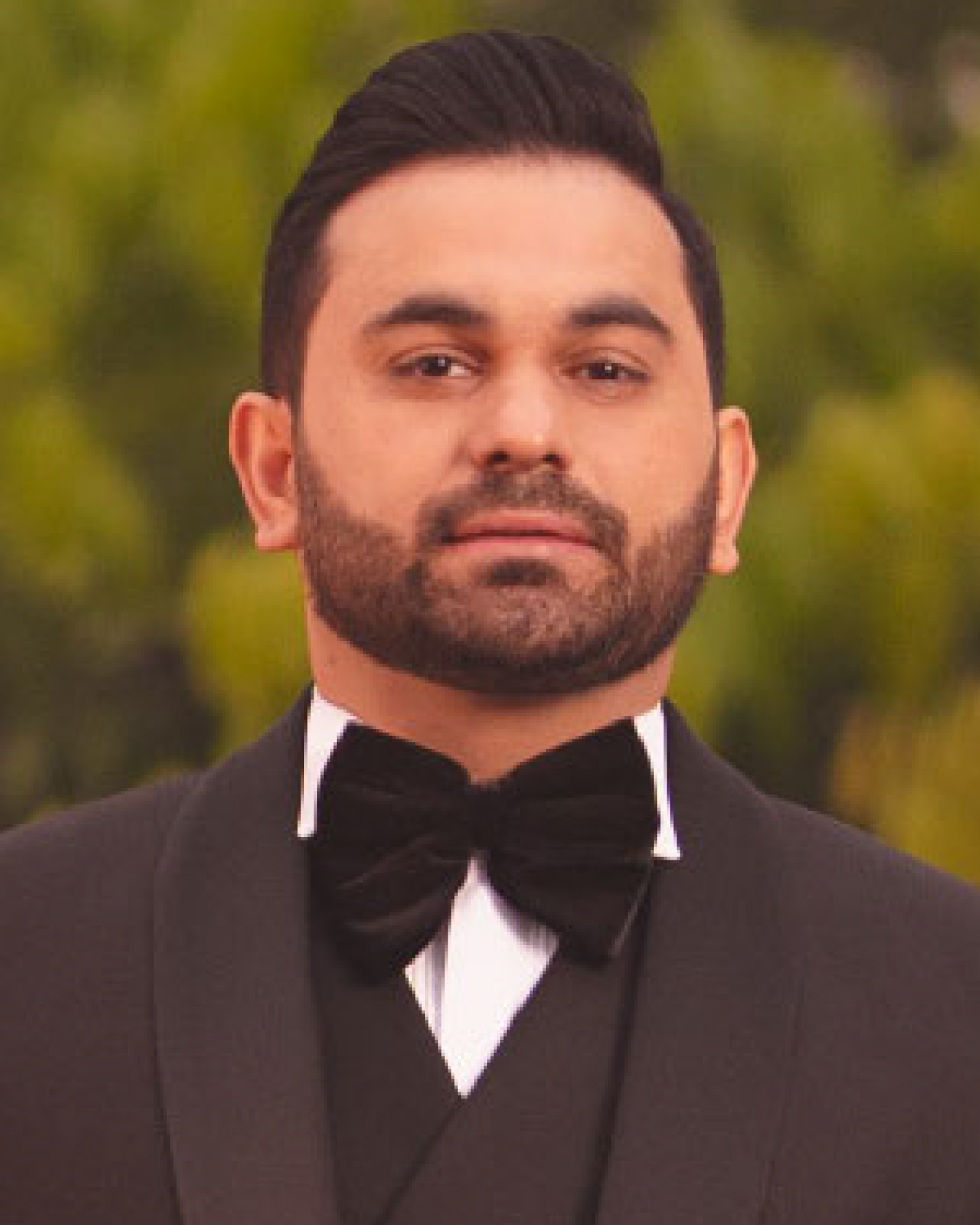}}]{Sajid Javed} received the B.Sc. degree in computer science from the University of Hertfordshire, Hatfield, U.K., in 2010, and the M.Sc. and Ph.D. degrees in computer science from Kyungpook National University, Daegu, Republic of Korea, in 2017.

He is currently a Faculty Member with Khalifa University of Science and Technology, Abu Dhabi, UAE. Prior to that, he was a Research Fellow with Khalifa University, from 2019 to 2021, and the University of Warwick, U.K., from 2017 to 2018. His research interests include visual object tracking in the wild, multi-object tracking, backgroundforeground modeling from video sequences, moving object detection from complex scenes, and cancer image analytics, including tissue phenotyping, nucleus detection, and nucleus classification problems. His research themes involve developing deep neural networks, subspace learning models, and graph neural networks.
\end{IEEEbiography}

\begin{IEEEbiography}[{\includegraphics[width=1in,height=1.25in,clip,keepaspectratio]{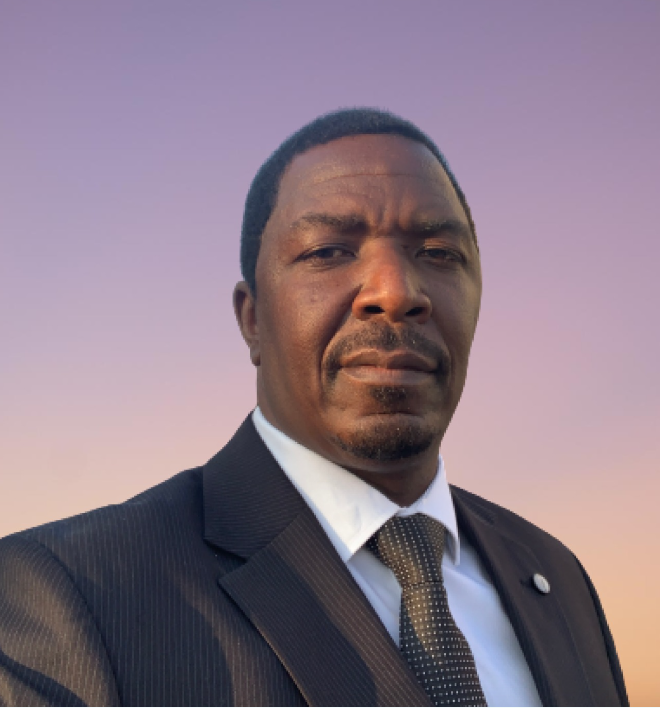}}]{Brain Moyo} received the Bachelor of Commerce degree in supply and operations management and the Bachelor of Technology degree
in quality management from the University of
South Africa, the Diploma degree in computer
studies from NCC Education, U.K., and the Cert
and Diploma degree in aeronautical engineering
from Zimbabwe. He received the Lean Six Sigma
Black Belt Certification. Throughout his career,
he has held various positions in the aviation industry, allowing him to develop a comprehensive understanding of the field. Currently, he is the Head of Research and Development with Sanad Aerotech A
Mubadala Company, where he is responsible for leading the development and
adoption of new technologies and innovations in the industry. His research
interests include bridging the gap between academia and industry to bring
emerging technologies to practical applications within the aviation industry.
He is a highly experienced aviation professional with over 27 years of
industry experience. He believes that the adoption of cutting-edge technology
can bring significant benefits to organizations and is passionate about helping
his company realize these benefits.
\end{IEEEbiography}

\begin{IEEEbiography}[{\includegraphics[width=1in,height=1.25in,clip,keepaspectratio]{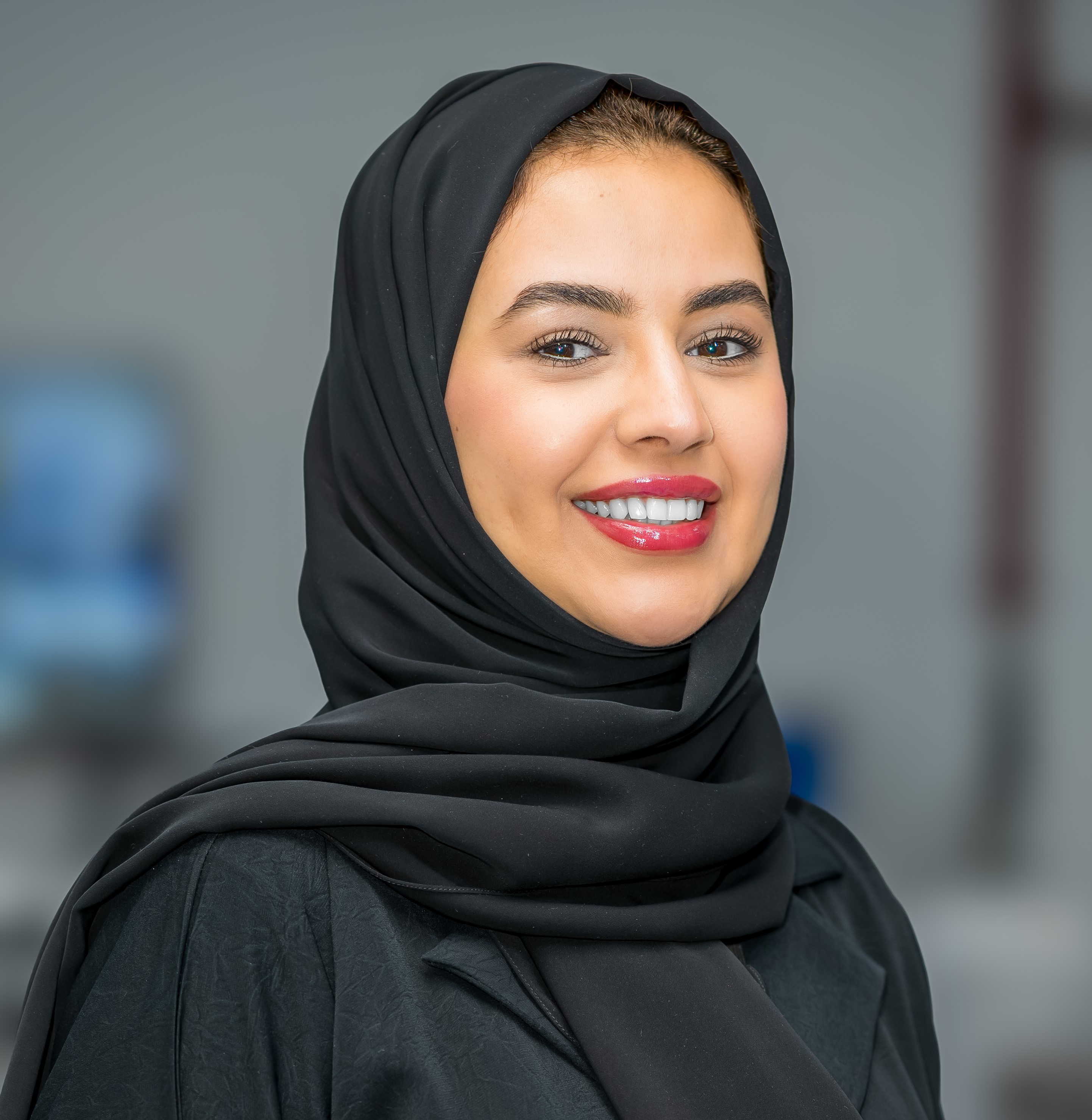}}]{Yusra Abdulrahman} (Member, IEEE)

received the B.Sc. degree from The University of
Arizona, in 2014, and the M.Sc. and Ph.D. degrees
from Massachusetts Institute of Technology and
Masdar Institute of Science and Technology
Cooperative Program (MIT and MICP), in
2016 and 2020, respectively. She is currently
an Assistant Professor with the Department of
Aerospace Engineering, Khalifa University. Her
expertise lies in robotics, artificial intelligence
(AI), and non-destructive testing (NDT). She has received awards from the
UAE Ministry of Energy and Industry for her research contributions.

\end{IEEEbiography}

\end{document}